\documentclass{article}

\usepackage{arxiv}
\usepackage{booktabs}
\usepackage[utf8]{inputenc} % allow utf-8 input
\usepackage[T1]{fontenc}    % use 8-bit T1 fonts
\usepackage{hyperref}       % hyperlinks
\usepackage{url}            % simple URL typesetting
\usepackage{booktabs}       % professional-quality tables
\usepackage{amsfonts}       % blackboard math symbols
\usepackage{nicefrac}       % compact symbols for 1/2, etc.
\usepackage{microtype}      % microtypography
\usepackage{lipsum}
\usepackage{fancyhdr}       % header
\usepackage{graphicx}       % graphics
\graphicspath{{media/}}     % organize your images and other figures under media/ folder
\usepackage{enumitem}
\usepackage{multirow}
\usepackage{amsmath}
\usepackage{pifont}
\newcommand{\cmark}{\ding{51}}%
\newcommand{\xmark}{\ding{55}}
\usepackage{xcolor}
\usepackage{verbatim}
\usepackage{adjustbox}

\newcommand{\EE}{\mathcal{E}}
\newcommand{\DD}{\mathcal{D}}
\newcommand{\TT}{\mathcal{T}}

\newcommand{\Loss}{\mathcal{L}}

\newcommand{\mr}[1]{\mathrm{#1}}

\newcommand{\ppm}{\,\scriptsize$\pm$}
\newcommand{\mypar}[1]{\vspace{2pt}\noindent\textbf{#1~}}

\usepackage{colortbl}

\title{FDS: Feedback-guided Domain Synthesis with Multi-Source Conditional Diffusion Models for Domain Generalization}

\author{Mehrdad Noori\thanks{Correspondence to \href{mailto:mehrdad.noori.1@ens.etsmtl.ca}{mehrdad.noori.1@ens.etsmtl.ca}} \And Milad Cheraghalikhani  \And Ali Bahri \And Gustavo A. Vargas Hakim \And David Osowiechi \And Moslem Yazdanpanah \And Ismail Ben Ayed \And Christian Desrosiers \AND \\ 
\'ETS Montreal, Canada \\ International Laboratory on Learning Systems (ILLS)}
\hypersetup{
pdftitle={main},
pdfsubject={cs.ML, cs.AI, cs.CV},
pdfauthor={Mehrdad Noori, Milad Cheraghalikhani, Ali Bahri, Gustavo A. Vargas Hakim, David Osowiechi, Moslem Yazdanpanah, Ismail Ben Ayed, Christian Desrosiers},
pdfkeywords={Domain Generalization, Diffusion Models, Domain Synthesis},
}

\begin{document}
\maketitle

\begin{abstract}
  %Standard deep learning architectures such as convolutional neural networks and vision transformers often fail to generalize to previously unseen domains due to the implicit assumption that both source and target data are drawn from independent and identically distributed (i.i.d.) populations. In response, 
  Domain Generalization techniques aim to enhance model robustness by simulating novel data distributions during training, typically through various augmentation or stylization strategies. However, these methods frequently suffer from limited control over the diversity of generated images and lack assurance that these images span distinct distributions. To address these challenges, we propose \textbf{FDS}, \textbf{F}eedback-guided \textbf{D}omain \textbf{S}ynthesis, a novel strategy that employs diffusion models to synthesize novel, pseudo-domains by training a single model on all source domains and performing domain mixing based on learned features. By incorporating images that pose classification challenges to models trained on original samples, alongside the original dataset, we ensure the generation of a training set that spans a broad distribution spectrum. Our comprehensive evaluations demonstrate that this methodology sets new benchmarks in domain generalization performance across a range of challenging datasets, effectively managing diverse types of domain shifts. The code can be found at: \url{https://github.com/Mehrdad-Noori/FDS}.
\end{abstract}

% keywords can be removed
% \keywords{First keyword \and Second keyword \and More}

\section{Introduction}

Deep learning architectures, including Convolutional Neural Networks (CNNs) and Vision Transformers (ViTs), have significantly advanced the field of computer vision, achieving state-of-art results in tasks like classification, semantic segmentation, and object detection. Despite these advancements, such models commonly operate under the simplistic assumption that training data (source domain) and post-deployment data (target domain) share identical distributions. This overlook of distributional shifts results in performance degradation when models are exposed to out-of-distribution (OOD) data~\cite{recht2019imagenet,hendrycks2019benchmarking}. Domain adaptation (DA)~\cite{lu2020stochastic,saito2018maximum} and Test-Time Adaptation (or Training)~\cite{wang2020tent, hakim2023clust3, osowiechi2023tttflow} strategies have been developed to mitigate this issue by adjusting models trained on the source domain to accommodate a predefined target domain. Nonetheless, these strategies are constrained by their dependence on accessible target domain data for adaptation, a prerequisite that is not always feasible in real-world applications. Furthermore, adapting models to each novel target domain entails considerable computational overhead, presenting a practical challenge to their widespread implementation.

Domain generalization (DG)~\cite{blanchard2011generalizing} aims to solve the issue of domain shift by training models using data from one or more source domains so that they perform well \emph{out-of-the-box} on new, unseen domains. Recently, various techniques have been developed to tackle this problem~\cite{zhou2022domain,wang2022generalizing}, including \emph{domain aligning}~\cite{hu2020domain,mahajan2021domain,li2020domain}, \emph{meta-learning}~\cite{li2018learning,balaji2018metareg}, \emph{data augmentation}~\cite{shi2020towards,volpi2018generalizing,shankar2018generalizing}, \emph{ensemble learning}\cite{zhou2021domain}, \emph{self-supervised learning}~\cite{carlucci2019domain,albuquerque2020improving}, and \emph{regularization methods}~\cite{huang2020self, cha2021swad}. These strategies are designed to make models more adaptable and capable of handling data that they were not explicitly trained on, making them more useful in real-world situations where the exact nature of future data cannot be predicted. Among these techniques, a notable category focuses on synthesizing samples from different distributions to mimic target distributions. This is achieved through strategies like \emph{image transformation}~\cite{shi2020towards, zhang2020generalizing}, \emph{style transfer}~\cite{somavarapu2020frustratingly, borlino2021rethinking}, \emph{learnable augmentation networks}~\cite{zhou2020deep, zhou2020learning, carlucci2019hallucinating}, and \emph{feature-level stylization}~\cite{zhou2021domain, noori2024tfs, cheraghalikhani2024structure}. However, many of these methods face challenges in controlling the synthesis process, often resulting in limited diversity where primarily only textures are altered.  

\begin{figure*}[t]
  \centering
  \includegraphics[width=0.95\textwidth]{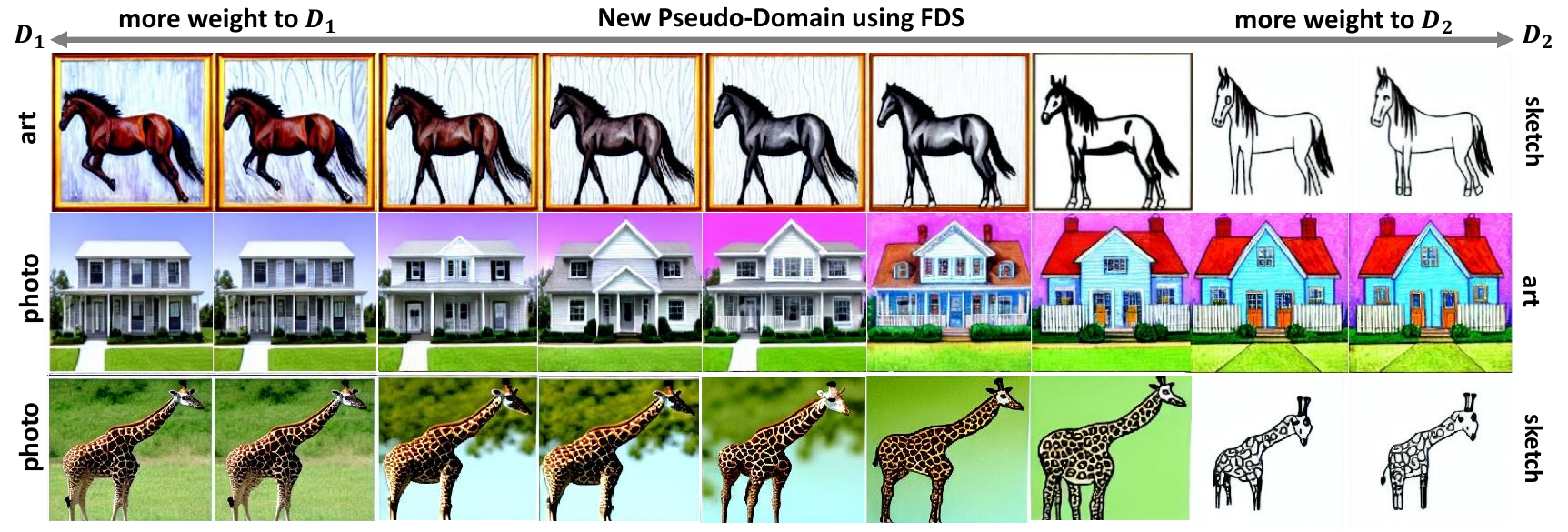}
  % \caption{Bridging Domain Gaps with FDS: Comprehensive Distribution Coverage Across Domains.}
  \caption{Generating new, pseudo-domains with FDS: Comprehensive distribution coverage from domain \(D_1\) to \(D_2\).}
  \vspace{-0.7cm}
  \label{fig:inter}
\end{figure*}

In this study, we introduce an innovative approach using diffusion model, named Feedback-guided Domain Synthesis (FDS), to address the challenge of domain generalization. Known for their exceptional ability to grasp intricate distributions and semantics, diffusion models excel at producing high-quality, realistic samples~\cite{dhariwal2021diffusion, ho2020denoising, rombach2022high, song2020denoising}. We exploit this strength by training a single diffusion model that is conditioned on various \emph{domains} and \emph{classes} present in the training dataset, aiming to master the distribution of source domains. 

As illustrated in Figure~\ref{fig:inter}, through the process of domain interpolation and mixing during generation, we create images that appear to originate from novel, pseudo-domains. To ensure the development of a robust classifier, we initially select generated samples that are difficult for a model trained solely on the original source domains to classify. Subsequently, we train the model using a combination of these challenging images and the original dataset. In this manner, we make sure that our model is fed with images of the widest possible diversity, thereby significantly enhancing its ability to generalize across unseen domains. Our contributions can be summarized as follows:
\begin{itemize}[leftmargin=*,labelsep=1.0em]
\item We introduce a novel DG approach that leverages diffusion models conditioned on multiple domains and classes to generate samples from novel, pseudo-domains through domain interpolation. This approach increases the diversity and realism of the generated images;
\item We propose an innovative strategy for selecting images that pose a challenge to a classifier trained on original images, ensuring the diversity of the final sample set. By incorporating these challenging images with the original dataset, we ensure a comprehensive and diverse training set, significantly improving the model's generalization capabilities;
\item We conduct extensive experiments across various benchmarks and perform different analyses to validate the effectiveness of our method. These experiments demonstrate FDS's ability to significantly improve the robustness and generalization of models across a wide range of unseen domains, achieving SOTA performance.
\end{itemize}

\section{Related Works}
\label{sec:related}

\subsection{Domain Generalization (DG)}
Domain Generalization, a concept first introduced by Blanchard et al. in 2011~\cite{blanchard2011generalizing}, has seen growing interest in the field of computer vision. This interest has spurred the development of a broad spectrum of methods aimed at enabling models to generalize across unseen domains. These include approaches based on \emph{domain alignment} like moment matching~\cite{peng2019moment}, discriminant analysis~\cite{hu2020domain} and domain-adversarial learning~\cite{li2018deep}, \emph{meta-learning} approaches~\cite{li2018learning,balaji2018metareg} which solve a bi-level optimization problem where the model is fine-tuned on meta-source domains to minimize the error on a meta-target domain, \emph{ensemble learning} techniques~\cite{zhou2021domain,mancini2018best} improving robustness to OOD data by training multiple models tailored to specific domains, \emph{self-supervised learning} methods fostering domain-agnostic representations through the pre-training of models on unsupervised tasks~\cite{carlucci2019domain,wang2020learning,ghifary2015domain} or via contrastive learning~\cite{kim2021selfreg}, approaches leveraging \emph{disentangled representation learning}~\cite{li2017deeper,chattopadhyay2020learning} segregating domain-specific features from those common across all domains, and \emph{regularization} methods which build on the empirical risk minimization (ERM) framework~\cite{Gulrajani2021InSO}, incorporating additional objectives such as distillation~\cite{wang2021embracing, sultana2022self}, stochastic weight averaging~\cite{cha2021swad}, or distributionally robust optimization (DRO)~\cite{sagawa2019distributionally} to promote generalization.

A key strategy in DG, \emph{data augmentation} aims to enhance model robustness against domain shifts encountered during deployment. This objective is pursued through a variety of techniques, such as learnable augmentation, off-the-shelf style transfer, and augmentation at the feature level. Learnable augmentation models utilizes networks to create images from training data, ensuring their distribution diverges from that of the source domains~\cite{zhou2020deep,carlucci2019hallucinating,zhou2020learning}. Meanwhile, off-the-shelf style transfer based methods seek to transform the appearance of images from one domain to another or modify their stylistic elements, often through Adaptive Instance Normalization (AdaIN)\cite{huang2017arbitrary,somavarapu2020frustratingly}. Unlike most augmentation approaches that modify pixel values, some propose altering features directly, a technique inspired by the finding that CNN features encapsulate style information\cite{mancini2020towards,zhou2021domain} in their statistics.

\subsection{Diffusion Models}

Diffusion models have recently surpassed Generative Adversarial Networks (GANs) as the leading technique for image synthesis. Innovations like denoising diffusion probabilistic models (DDPMs)~\cite{ho2020denoising} and denoising diffusion implicit models (DDIMs)~\cite{song2020denoising} have significantly sped up the image generation process. Rombach et al.~\cite{rombach2022high} introduced latent diffusion models (LDMs), also known as stable diffusion models, which enhance both training and inference efficiency while facilitating text-to-image and image-to-image conversions. Extensions of these models, such as stable diffusion XL (SDXL)~\cite{podell2023sdxl} and ControlNet~\cite{zhang2023adding}, have been developed to further guide the generation process with additional inputs, such as depth or semantic information, allowing for more controlled and versatile image creation. Recent studies have shown that augmenting datasets using diffusion models can improve performance in general vision tasks~\cite{azizi2023synthetic, dunlap2023diversify}.

Despite the capabilities of diffusion models in generating images, their application in domain generalization has been minimally explored.
Miao et al.~\cite{miao2024domaindiff} introduced DomainDiff, a method that boosts OOD generalization by training a Word-to-Image Mapping (WIM) with diffusion models to generate additional synthetic data.
%, enhancing performance on unseen domains.
Yue et al.~\cite{yu2023distribution} proposed a method called DSI to address OOD prediction by transforming testing samples back to the training distribution using multiple diffusion models, each trained on a single source distribution. While effective, DSI requires multiple models during prediction, making it impractical for real-time deployment. In contrast, our method only uses diffusion during training to synthesize pseudo-domains and select more challenging images, keeping the computational complexity the same during testing while achieving significantly better performance.

In another study, CDGA~\cite{hemati2023cross} uses diffusion models to generate synthetic images that fill the gap between different domain pairs through a simple interpolation. While effective, CDGA relies on generating a large number of synthetic images (e.g., 5M for PACS) and employs naive interpolation. Additionally, CDGA generates samples with additional text descriptions that are not provided in standard DG benchmarks. In contrast, our method employs novel and more efficient interpolation techniques and a unique filtering mechanism that selects challenging images. We demonstrate that both of these innovations are crucial for improving generalization.

\begin{figure*}[t]
  \centering
  \includegraphics[width=0.95\textwidth]{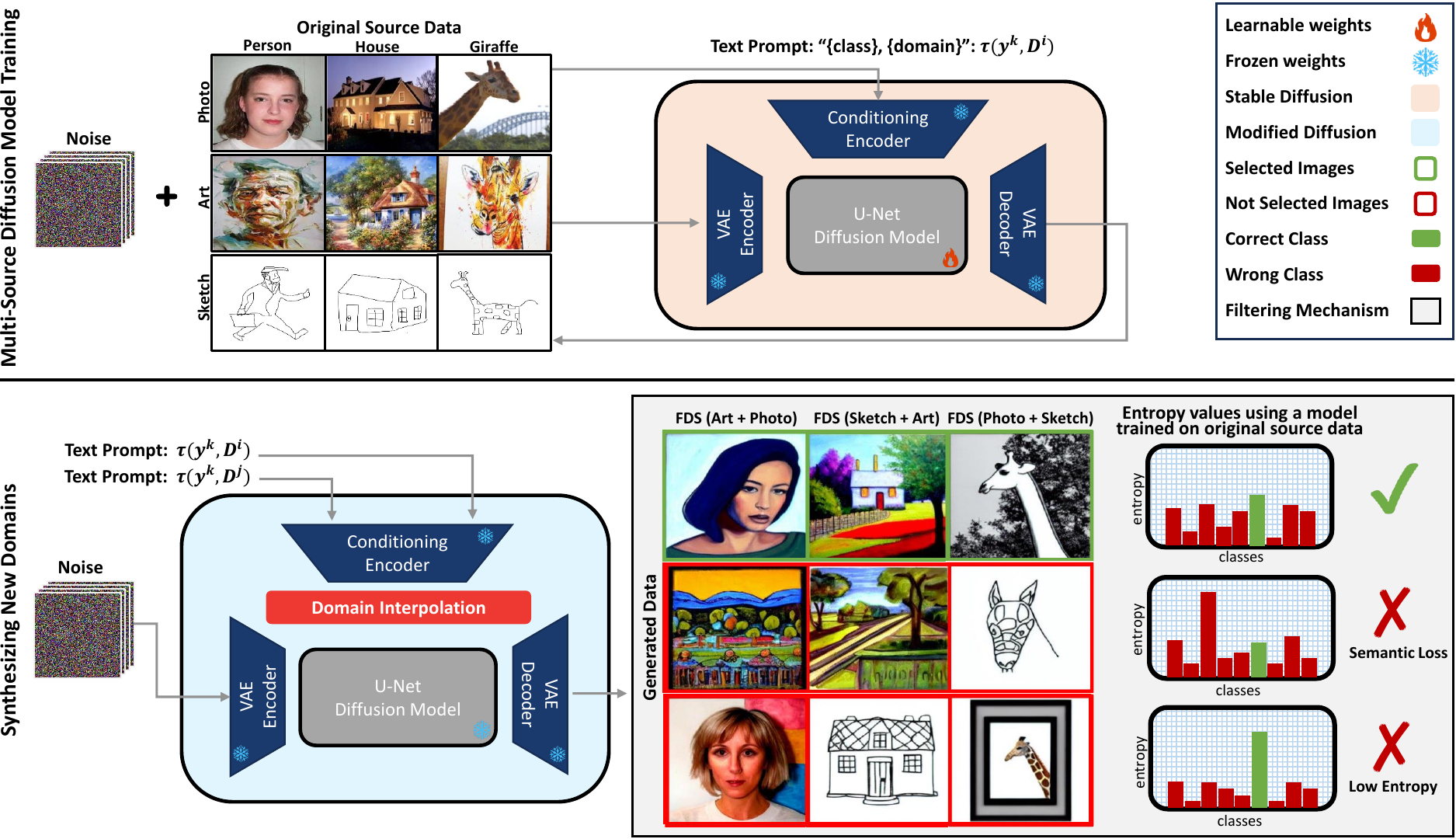}
  \caption{Overview of the proposed architecture for FDS. (top) Multi-source training of diffusion model conditioned on class and domain of the training images. (bottom) Generating novel pseudo-domain using the proposed interpolation and filtering mechanism of FDS.}
  \label{fig:arch}
\end{figure*}

\section{Theoretical Motivation}
Our classifier, represented by $f$ with parameters $\theta$, aims to craft a unified model from $n$ source domains $\{\DD^1, \ldots, \DD^n\}$ that adapts to a novel target domain $\DD^\TT$. Within any domain $\DD$, we measure classification loss by
\begin{equation}
\mathcal{L}_{\DD}(\theta) \, = \, \mathbb{E}_{(x,y)\sim\DD}\,\big[\ell(f(x;\theta), y)\big],
\end{equation}
where $x$ and $y$ denote the input and its corresponding label, respectively, and $\ell(f(x; \theta), y)$ is the cross entropy loss in this work.

Empirical Risk Minimization (ERM)~\cite{sain1996nature} forms the foundation for training our models, aiming to reduce the mean loss across training domains:
\begin{equation}
\min_{\theta} \ \sum_{i} \frac{1}{|\DD^i|} \sum_{k=1}^{|\DD^i|} \ell(f(x_i^k; \theta), y_i^k)
\end{equation}
Here, $|\DD^i|$ counts the number of samples in domain $i$, with $x_i^k$ as the $k$-th sample and $y_i^k$ its label. However, the accuracy of ERM-trained models drops when data shifts occur across domains due to inadequate OOD generalization. To enhance ERM, Chapelle et al. introduced Vicinal Risk Minimization (VRM)~\cite{chapelle2000vicinal}, which substitutes point-wise estimates with density estimation in the vicinity distribution around every observation within each domain. Practical implementation often involves data augmentation to introduce synthetic samples from these density estimates. Traditional augmentation processes a single data point $x_i^k$ from domain $i$ to yield $\tilde{x}_i^k = g(x_i^k)$, where $g(\cdot)$ denotes a basic transformation and $\tilde{x}_i^{k}$ is the modified data point.

Despite the potential of VRM to boost performance on OOD samples, it does not completely bridge domain gaps. The root cause lies in the inability of ERM methods to anticipate shifts in data distribution, which simple augmentation within domains does not address.  Hence, domain generalization requires more robust transformations to extend the model's applicability beyond training domains. Muller et al.~\cite{muller2021trivialaugment} have demonstrated that a wider range of transformations outperforms standard methods. However, Aminbeidokhti et al.~\cite{aminbeidokhti2024domain} advise that aggressive augmentations could distort the essential characteristics of images, pointing to the necessity of a mechanism to filter out extreme alterations. The emergence of diffusion models, proficient in reproducing diverse data distributions, facilitates such advanced sampling approaches. Our goal, therefore, is to generate images that span across domain gaps, lessen the variability between data distributions, and introduce a system to exclude trivial or excessive modifications.

\section{Method}
\label{sec:method}

To advance the generalization ability of our classifier, we aim to utilize generated image samples that traverse domain gaps yet retain class semantics. This objective is realized through a three-step methodology at the core of our FDS approach. \textbf{Step 1:} we begin by training an image generator that masters the class-specific distributions of all source domains, enabling the synthesis of class-consistent samples within those domains. \textbf{Step 2:} we then introduce a strategy for generating inter-domain images to span the domain gaps effectively. \textbf{Step 3:} last, we filter overly simplistic samples from synthetic inter-domain images and train the final model on this refined set alongside original domain images for enhanced OOD generalization. The process is illustrated in Figure~\ref{fig:arch}.

%In the follow subsections, we present each of these steps in detail.

\subsection{Image Generator}

Diffusion Models are designed to approximate the data distribution $p(x)$ by reversing a predefined Markov Chain of $T$ steps, effectively denoising a sample in stages. These stages are modeled as a sequence of denoising autoencoder applications $e_{\theta}(x_t, t)$, for $t = 1 \ldots T$, which gradually restore their input $x_t$. This iterative restoration is formalized by the following objective:
\begin{equation}
\Loss_{DM} \, = \, \mathbb{E}_{x,\epsilon\sim\mathcal{N}(0,1),\,t}\left[ \| \epsilon - \epsilon_{\theta}(x_t, t) \|^2 \right],
\end{equation}
where $t$ is drawn uniformly from $\{1, \ldots, T\}$. Utilizing perceptual compression models, encoded by $\EE$ and decoded by $\mathcal{\tilde{E}}$, we access a latent space that filters out non-essential high-frequency details. By placing the diffusion process in this compressed space, our objective is then reformulated as
\begin{equation}
\Loss_{LDM} \, = \, \mathbb{E}_{\EE(x), \epsilon \sim \mathcal{N}(0,1),\,t}\left[\left\| \epsilon - \epsilon_{\theta}(z_t, t) \right\|^2 \right].
\end{equation}

Diffusion models can also handle conditional distributions $p(z|c)$. To enable this conditioning, building upon the work of Rombach et al. \cite{rombach2022high}, we utilize a condition encoder $\tau_{\theta}$ that maps $c$ onto an intermediate representational space $\tau(c) \in \mathbb{R}^{M \times d_{\tau}}$. Following Stable Diffusion, we employ the CLIP-tokenizer and implement $\tau$ as a transformer to infer a latent code. This representation is subsequently integrated into the UNet using a cross-attention mechanism, culminating in our final enhanced objective:
\begin{equation}
\Loss_{LDM}^{\mr{cond}} \, = \, \mathbb{E}_{\EE(x),\,c,\,\epsilon\sim\mathcal{N}(0,1),\,t}\left[ \left\| \epsilon - \epsilon_{\theta}(z_t, t, \tau(c)) \right\|^2 \right], 
\end{equation}

This approach enables a single diffusion model to understand and generate images across different domains for each class, using text-based conditions (prompts). By dynamically adjusting conditions, the model efficiently learns varied representations without needing multiple models for each domain. Thus, we employ a textual template $c(x)$, denoting ``$\texttt{[y]}$, $\texttt{[D]}$'' where $\texttt{y}$ is the class label and $\texttt{D}$ is the domain name of the input $x$, and proceed to train our diffusion model using this template across all images from the source domains. Upon completing training, we can create a new sample for class $y^k$, belonging to the set $\{y^1, \ldots, y^m\}$, within domain $\DD^i$. This is achieved by decoding a denoised representation after $t$ timesteps, $\tilde{x}_t^{i,k} =  \tilde{\EE} (\Phi_t (\DD^i, y^k))$, where $\Phi_t (\DD^i, y^k)$ is the denoised representation that originates from random Gaussian noise conditioned on $\DD^i$ and $y^k$.

\subsection{Domain Mixing}
We propose two mixing strategies to synthesize images from new, pseudo distributions, based on \emph{noise-level interpolation} and \emph{condition-level interpolation}.

\subsubsection{Noise Level Interpolation}

Consider our dataset comprising $n$ source domains $\{\DD^1, \ldots, \DD^n\}$ and target classes $y^k$ from the set $\{y^1, \ldots, y^m\}$. Utilizing a trained image generator that initiates with random Gaussian noise, we denoise this input over $t$ steps to produce a synthetic, denoised representation $\tilde{z}_t^{i,k} = \Phi_t (\DD^i, y^k)$ indicative of domain $\DD^i$ and class $y^k$. To synthesize a sample that merges the characteristics of domains $\DD^i$ and $\DD^j$ for class $y^k$, we employ a single diffusion model, conditioned dynamically to capture the essence of both domains. This process aims to generate a sample that embodies the transitional features between these domains, effectively bridging the domain gap.

The model begins its process from the same initial random Gaussian noise, adapting its denoising trajectory under two distinct conditions, $\tau (\DD^i, y^k )$ and $\tau (\DD^j, y^k)$, up to a specific timestep $T$. This dual-conditioned approach ensures that the evolving representation up to $T$ incorporates influences from both domains, guided by the respective conditional inputs.

From timestep $T$ onwards, until the final representation is formed, the model blends the outputs from these dual paths at each step. Specifically, for each step $t > T$, we form a mixed representation $\tilde{z}_{t}$ as:
\begin{equation}
\tilde{z}_{t} \, = \, \alpha\,\Phi_{t} (\DD^i, y^k) \, + \, (1\!-\!\alpha)\,\Phi_{t} (\DD^j, y^k),
\end{equation}
where $\alpha$ is a predefined mixing coefficient that dictates the blend of domain characteristics in the output. This combined representation $\tilde{z}_{t}$ is then used as the basis for the model's next denoising step, integrating features from both $\DD^i$ and $\DD^j$ for class $y^k$. Through this iterative mixing, the model ensures a gradual and cohesive fusion of domain-specific attributes, leading to a synthesized sample that seamlessly spans the gap between the domains for class $y^k$.

\vspace*{-3pt}
\subsubsection{Condition Level Interpolation}
We also propose Condition Level Interpolation to generate images that effectively bridge domain gaps. This technique relies on manipulating the conditions fed into our diffusion model to guide the synthesis of new samples. Specifically, for a target class $y^k$ and two distinct domains, $\DD^i$ and $\DD^j$, we employ our encoder $\tau(c)$ to create separate condition representations for each domain-class pair: $(y^k, \DD^i)$ and $(y^k, \DD^j)$. 

The core of this strategy involves blending these condition representations using a mixing coefficient $\alpha$, leading to a unified condition:
\begin{equation}
c_{\mr{mixed}} \, = \, \alpha\,\tau(c_{y^k}, \DD^i) \, + \, (1\!-\!\alpha)\,\tau(c_{y^k}, \DD^j).
\end{equation}
This mixed condition $c_{\text{mixed}}$ then orchestrates the generation process from the initial step, ensuring that the diffusion model is consistently influenced by attributes from both domains. By initiating this conditioned blending from the beginning of the diffusion process, we ensure a harmonious integration of domain characteristics throughout the generation of the synthetic image.

\subsection{Filtering Mechanism}

Through our mixing strategies, we create synthetic samples $\tilde{x}_{i,j}^k$ that not only synthesize class $y^k$ traits but also blend features from domains $\DD^i$ and $\DD^j$, thereby aiming to bridge the domain gaps. This approach generates a synthetic dataset comprising $\tilde{N}$ samples for each combination of class index $k$ and domain index pair $(i,j)$, structured as $S_{i,j}^k = \{\tilde{x}_{i,j}^{k,(r)} \, | \, 1 \leq i < j \leq n, \,\, 1 \leq k \leq m, \,\, 1 \leq r \leq \tilde{N}\}$, ensuring diversity via distinct random Gaussian noise initiation for each sample.

The utility of these synthetic samples in improving model generalization varies, prompting an entropy-based evaluation to identify those with the greatest potential. High entropy scores, indicating prediction uncertainty by a classifier $h(x)$ trained on the original dataset, suggest that such samples may come from previously unseen distributions. This characteristic posits these high-entropy samples as prime candidates for training, hypothesized to challenge the classifier significantly and aid in covering the domain gaps. Further refining this selection, we only include samples correctly predicted as their target class by $h(x)$, ensuring the exclusion of samples that have lost semantic integrity during the diffusion process.

Let $\mathcal{C}_{i,j}^k$ be the subset of samples in $S_{i,j}^k$ which are correctly classified by $h(x)$:
\begin{equation}
\mathcal{C}_{i,j}^k \, = \, \big\{ \tilde{x}_{i,j}^{k,(r)} \, | \, 
h(\tilde{x}_{i,j}^{k,(r)}) = y^k \big\}.
\end{equation}
We choose from $\mathcal{C}_{i,j}^k$ the $N_L$ samples with highest entropy (the entropy of a $k$-class discrete probability distribution $p$ is given by $\mathcal{H}(p) = -\sum_k p_k \log p_k$) to form a set of selected samples $\tilde{\DD}_{i,j}^k$. Last, we combine the synthetic samples, created for each class and domain pair, to the original dataset $\mathcal{O}$ to obtain the final augmented training set
\begin{equation}
\mathcal{A} \, = \, \mathcal{O} \, \cup \, \big\{\tilde{\DD}_{i,j}^k \, | \, 1 \leq i < j \leq n, \,\, 1 \leq k \leq m\big\}
\end{equation}
Training the final classifier on $\mathcal{A}$ not only enriches the dataset but also ensures robust model generalization across diverse domain landscapes.

\section{Experimental Setup}
\label{sec:exp}
\vspace*{-2pt}

\begin{table}[t]
    \centering
    \small
    \tabcolsep=0.05cm
    \renewcommand{\arraystretch}{1} % Increase space between rows
    \adjustbox{max width=\linewidth}{
    \begin{tabular}{ll|c|p{0.1cm}cp{0.1cm}cp{0.1cm}cp{0.1cm}|cp{0.1cm}c}
    \toprule
     & \textbf{Method} & \textbf{ Aug. }           && \textbf{PACS}   && \textbf{VLCS}             && \textbf{Office}          && \textbf{  ~Avg. }              \\
    \midrule
    \multirow{15}{*}{\rotatebox{90}{Standard Methods}}~ 
    & ERM (\emph{baseline})~\cite{Gulrajani2021InSO} & \xmark  && 85.5\ppm0.2              && 77.5\ppm0.4              && 66.5\ppm0.3            && 76.5 \\
    & ERM (\emph{reproduced}) & \xmark  && 84.3\ppm1.1              && 76.2\ppm1.1             &&64.6\ppm1.1           &&75.0  \\
    & IRM~\cite{arjovsky2019invariant} & \xmark           && 83.5\ppm0.8              && 78.5\ppm0.5              && 64.3\ppm2.2            && 75.4 \\
    & GroupDRO~\cite{sagawa2019distributionally} & \xmark && 84.4\ppm0.8              && 76.7\ppm0.6              && 66.0\ppm0.7            && 75.7 \\
    & Mixup~\cite{yan2020improve} & \cmark             && 84.6\ppm0.6              && 77.4\ppm0.6              && 68.1\ppm0.3            && 76.7 \\
    & CORAL~\cite{sun2016deep} & \xmark                  && 86.2\ppm0.3              && 78.8\ppm0.6              && 68.7\ppm0.3            && 77.9 \\
    & MMD~\cite{li2018domain} & \xmark                   && 84.6\ppm0.5              && 77.5\ppm0.9              && 66.3\ppm0.1            && 76.1 \\
    & DANN~\cite{ganin2016domain} & \xmark                && 83.6\ppm0.4              && 78.6\ppm0.4              && 65.9\ppm0.6            && 76.0 \\
    & SagNet~\cite{Nam_2021_CVPR} & \cmark              && 86.3\ppm0.2              && 77.8\ppm0.5              && 68.1\ppm0.1            && 77.4 \\
    & RSC~\cite{huang2020self} & \cmark                && 85.2\ppm0.9              && 77.1\ppm0.5              && 65.5\ppm0.9            && 75.9 \\
    & Mixstyle~\cite{zhou2021domain} & \cmark            && 85.2\ppm0.3              && 77.9\ppm0.5              && 60.4\ppm0.3            && 74.5 \\
    & mDSDI~\cite{bui2021exploiting} & \xmark             && 86.2\ppm0.2              && 79.0\ppm0.3              && \underline{69.2\ppm0.4}            && 78.1 \\
    & SelfReg~\cite{kim2021selfreg} & \cmark              && 85.6\ppm0.4             &&  77.8\ppm0.9              && 67.9\ppm0.7            && 77.1 \\
    & DCAug~\cite{aminbeidokhti2024domain} & \cmark      && 86.1\ppm0.7              && 78.6\ppm0.4              && 68.3\ppm0.4            && 77.7 \\
    & DomainDiff~\cite{miao2024domaindiff} & \cmark      && 85.6\ppm0.6             && --              && 63.7\ppm0.6            && -- \\

    & DSI~\cite{yu2023distribution} & \cmark      && 86.9\ppm1.4             && --             && --            && -- \\

    & CDGA~\cite{hemati2023cross} & \cmark              && \underline{88.5\ppm0.5}             && \underline{79.6\ppm0.3}              && 68.2\ppm0.6            && \underline{78.8}  \\
    \rowcolor{gray!15}
    & \textbf{ERM + FDS (ours)} & \cmark   && \textbf{88.8\ppm0.1}               && \textbf{79.8\ppm0.5}             &&\textbf{71.1\ppm0.1}           &&\textbf{79.9}  \\
    \midrule
    \multirow{7}{*}{\rotatebox{90}{WA Methods}}~ 
    & SWAD (\emph{baseline})~\cite{cha2021swad} & \xmark                     && 88.1\ppm0.1              && \underline{79.1\ppm0.1}             && 70.6\ppm0.2            && 79.3\\
    & SWAD (\emph{reproduced}) & \xmark && 88.1\ppm0.4             && 78.9\ppm0.5             && 70.3\ppm0.4           && 79.1 \\
    & SelfReg SWA~\cite{kim2021selfreg} & \cmark              && 86.5\ppm0.3              && 77.5\ppm0.0              && 69.4\ppm0.2            && 77.8 \\
    & DNA~\cite{chu2022dna} & \xmark                       && 88.4\ppm0.1              && 79.0\ppm0.1              && \underline{71.2\ppm0.1}            && 79.5 \\
    & DIWA~\cite{rame2022diverse} & \cmark                 && \underline{88.8\ppm0.4}              && 79.1\ppm0.2              && 71.0\ppm0.1            && \underline{79.6} \\
    & TeachDCAug~\cite{aminbeidokhti2024domain} & \cmark   && 88.4\ppm0.2              && 78.8\ppm0.4              && 70.4\ppm0.2            && 79.2 \\
    \rowcolor{gray!15}
    & \textbf{SWAD + FDS (ours)} & \cmark                          && \textbf{90.5\ppm0.3}    &&  \textbf{79.7\ppm0.5}   &&  \textbf{73.5\ppm0.4} && \textbf{81.3} \\
    \bottomrule
    \end{tabular}}
    \vspace{0.2cm}
    \caption{Leave-one-out accuracy\,(\%) results on the PACS, VLCS, and OfficeHome benchmarks. ``\textbf{Aug.}'' indicates whether advanced augmentation or domain mixing techniques are used. The \textbf{best results} and \underline{second-best results} are highlighted.}
    \label{tab:sota}
\end{table}

\mypar{Datasets.} Following~\cite{Gulrajani2021InSO}, we compare our proposed approach to the current state-of-art using three challenging datasets--\textbf{12 individual target domains}--with different characteristics: \texttt{PACS}~\cite{li2017deeper}, \texttt{VLCS}~\cite{fang2013unbiased}, and \texttt{OfficeHome}~\cite{venkateswara2017deep}. The PACS dataset has a total of 9,991 photos divided into four distinct domains, $d \in$ \{Art, Cartoon, Photo, Sketch\}, and seven distinct classes. The second dataset, \texttt{VLCS}, comprises 10,729 photos from four separate domains, $d \in$ \{Caltech101, LabelMe , SUN09, VOC2007\}, and five different classes. The third dataset, \texttt{OfficeHome}, includes a total of 15,588 photos taken from four domains, $d \in$ \{Art, Clipart, Product, Real\}, and 65 classes.

\mypar{Implementation Details.} To ensure a fair comparison, we adopt the DomainBed framework~\cite{Gulrajani2021InSO}, a comprehensive benchmark that encompasses prominent domain generalization (DG) methodologies under a uniform evaluation protocol. Following this framework, we employ a leave-one-out strategy for DG dataset assessment where one domain serves as the test set while the others form the training set. A subset of the training data, constituting 20\%, is designated as the validation set\footnote{\ The model with peak accuracy on this validation set is selected for evaluation on the test domain, providing unseen domain accuracy.}. The aggregate result for each dataset represents the mean accuracy derived from varying the test domain. To ensure reliability, experiments are replicated three times, each with a unique seed. Moreover, to rigorously test our method, we try both ERM and SWAD classifiers with FDS. The former is considered a baseline in standard training methods, while the latter serves as a baseline for weight averaging (WA) methods. SWAD is essentially ERM but with weight averaging applied during training using multiple steps based on the validation set. To have a fair comparison with other methods, for the ERM baseline, we strictly follow the hyperparameter tuning proposed in~\cite{cha2021swad}. For SWAD, as in the original work, we did not tune any parameters and used the default values. For image synthesis, the original Stable Diffusion framework~\cite{rombach2022high} is utilized with $\mr{DDIM}\!=\!50$ steps. The PACS and VLCS datasets prompt the generation of $N\!=\!32,000$ samples per class, whereas OfficeHome, with its 65 classes, necessitates $N \!=\! 16,000$ samples. Image generation spans an interpolation range of $\alpha\!\in\![0.3, 0.7]$ and a Noise Level Interpolation range of $\mathcal{T}\!\in\![20, 45]$. This diversified parameter selection, rather than optimizing hyperparameters per dataset, acknowledges each domain's unique shift. Our filtering mechanism then identifies the most informative images from the generated pool. The impact of dataset size, $N_L$, is studied further in Section~\ref{subsec:furthur}.

\section{Results}
\label{sec:res}
We first compare the performance of our FDS approach against SOTA DG methods across three benchmarks. We then present a detailed analysis investigating several key aspects of our approach, including the effectiveness of each proposed component, a comparison of different mixing strategies, its regularization capabilities, domain diversity visualization and quantification, the impact of data size, the efficacy of our filtering mechanism, and stability analysis during training. Additional analysis, visualizations, and detailed tables can be found in the Supplementary Material.

\subsection{Comparison with the State-of-the-art}

In Table~\ref{tab:sota}, we compare our approach with recent methods for domain generalization as outlined in the DomainBed framework~\cite{Gulrajani2021InSO}. Our method, when added to the baseline ERM classifier, shows an impressive improved accuracy of 4.5\% on the PACS dataset using the ResNet-50 model. On the VLCS dataset, our method sees a 3.6\% increase in accuracy over the ERM. For the OfficeHome dataset, our method outperforms the ERM baseline by 6.5\%. To test the strength of our method, we also applied it to SWAD, which is a the baseline for weight averaging (WA) domain generalization methods. Here, our method improves the performance by 2.4\%, 0.8\%, and 3.2\% on the PACS, VLCS, and OfficeHome datasets, respectively. 

Furthermore, when comparing with previous SOTA methods, our method outperforms CDGA~\cite{hemati2023cross} by 0.3\% on PACS, 0.2\% on VLCS, and 2.9\% on OfficeHome. In the context of weight averaging methods, our approach surpasses DIWA~\cite{rame2022diverse}, which trains multiple independent models, by 1.7\% on PACS, 0.6\% on VLCS, and 2.5\% on OfficeHome, setting a new benchmark for domain generalization. Please refer to the Supplementary Material for the full, detailed results for each dataset and its domains.

\begin{table}[t]
    \centering
    \tabcolsep=0.2cm
    \adjustbox{max width=\linewidth}{
      \begin{tabular}{lcccc|c}
        \toprule
        \multirow[b]{2}{*}{\bf Module} & \multicolumn{5}{c}{\bf Target Domains} \\
        \cmidrule(l{4pt}r{4pt}){2-6}
         &  \bf Art & \bf Cartoon & \bf Photo & \bf Sketch & \bf Avg.\\
        \midrule
        Baseline (SWAD~\cite{cha2021swad})      & 89.49\ppm0.2 & 83.65\ppm0.4 & 97.25\ppm0.2 & 82.06\ppm1.0 & 88.11\ppm0.4 \\
        + Basic Gen.   & 89.87\ppm0.1 & 85.59\ppm0.6 & 97.50\ppm0.3 & 83.07\ppm0.4 & 89.01\ppm0.4 \\
        + Interpolation                   & 91.38\ppm0.2 & 85.20\ppm0.6 & 97.73\ppm0.1 & 84.27\ppm0.9 & 89.65\ppm0.4 \\
        + Filtering                   & \textbf{91.80\ppm0.3} & \textbf{86.03\ppm0.8} & \textbf{98.05\ppm0.2} & \textbf{86.11\ppm0.1} & \textbf{90.50\ppm0.3}\\
      \bottomrule
    \end{tabular}
}
\vspace{0.2cm}
\caption{Comparative analysis of FDS component effects on accuracy (\%) across PACS dataset domains. ``\textbf{Basic Gen.}'' refers to generation without interpolation or filtering.}
\label{tab:component}
\end{table}

\subsection{Further Analysis}
\label{subsec:furthur}
For this section, we use the SWAD baseline to analyze the performance of our proposed method FDS, due to its stable performance as a WA method. Final settings are applied to the ERM baseline. All analyses in this section use the PACS dataset and the SWAD baseline unless otherwise stated.

\mypar{Ablation Study on Different Components.}
To evaluate the impact of each component on generalization, we systematically introduce each module and observe the enhancements. As detailed in Table~\ref{tab:component}, first incorporating domain-specific synthetic samples, generated without interpolation or filtering (Basic Gen.), yields a $0.9\%$ gain over the baseline. This increment validates our diffusion model's proficiency in capturing and replicating the class-specific distributions within domains, thus refining OOD performance through enhanced density estimation near original samples. Subsequently, integrating images from pseudo-novel domains via our interpolation strategy leads to a further $0.5\%$ enhancement, underscoring the mechanism's effectiveness in connecting distinct domains. Lastly, applying our entropy-based filtering to eliminate overly simplistic images and images with diminished semantic relevance results in a significant $2.4\%$ improvement against the SWAD model, a robust benchmark. These outcomes collectively underscore the efficacy of our approach in improving OOD generalization.

\begin{table}[t]
    \vspace{-0.2cm}
    \centering
    \tabcolsep=0.2cm
    \adjustbox{max width=\linewidth}{
      \begin{tabular}{lcccc|c}
        \toprule
        % & \multicolumn{5}{|c}{Accuracy\,(\%)} \\
        %Data Size (\#imgs/class) &  Art & Cartoon & Photo & Sketch & Avg.\\
        \multirow[b]{2}{*}{\bf Strategy} & \multicolumn{5}{c}{\bf Target Domains} \\
        \cmidrule(l{4pt}r{4pt}){2-6}
         &  \bf Art & \bf Cartoon & \bf Photo & \bf Sketch & \bf Avg.\\
        \midrule
        Baseline (SWAD~\cite{cha2021swad})                   & 89.49\ppm0.2 & 83.65\ppm0.4 & 97.25\ppm0.2 & 82.06\ppm1.0 & 88.11\ppm0.4 \\
        Noise Level Interpol.   & \textbf{91.95\ppm0.3} & 83.23\ppm0.1 & 97.56\ppm0.1 & 85.85\ppm0.3 & 89.65\ppm0.2 \\
        Condition Level Interpol.    & 91.80\ppm0.3 & \textbf{86.03\ppm0.8} & \textbf{98.05\ppm0.2} & \textbf{86.11\ppm0.1} & \textbf{90.50\ppm0.3} \\
        Both    & 91.13\ppm0.4 & 82.98\ppm0.2 & 97.90\ppm0.2 & 85.62\ppm0.8 & 89.41\ppm0.4 \\

      \bottomrule
    \end{tabular}
}
\vspace{0.2cm}
\caption{Impact of different interpolation strategies of FDS on PACS accuracy (\%).}
\label{tab:mixing}
\end{table}

\mypar{Mixing Strategies.} Our ablation study, presented in Table~\ref{tab:mixing}, reveals that both Noise Level Interpolation and Condition Level Interpolation significantly outperform the baseline, yet their effectiveness varies with the dataset's attributes and the nature of the domain shift. Noise Level Interpolation is optimal for minimal domain shifts, focusing on adjusting the noise aspect to bridge domain gaps. However, it falls short in scenarios with substantial domain differences, such as the transition to Cartoon or Sketch, where the source and target domains diverge significantly. In these cases, Condition Level Interpolation proves more advantageous, offering a robust mechanism for navigating complex domain shifts by manipulating higher-level semantic representations. When applying both methods simultaneously, the performance did not improve compared to using Condition Level Interpolation alone, possibly due to the increased complexity. Nonetheless, it still performed better than the baseline. Based on these results, we use Condition Level Interpolation as our final interpolation method.

\begin{figure}[tb]
    \centering    
    %\dorowcolors   
    \begin{minipage}[t]{0.485\linewidth}
    \centering
      \includegraphics[width=1.0\textwidth]{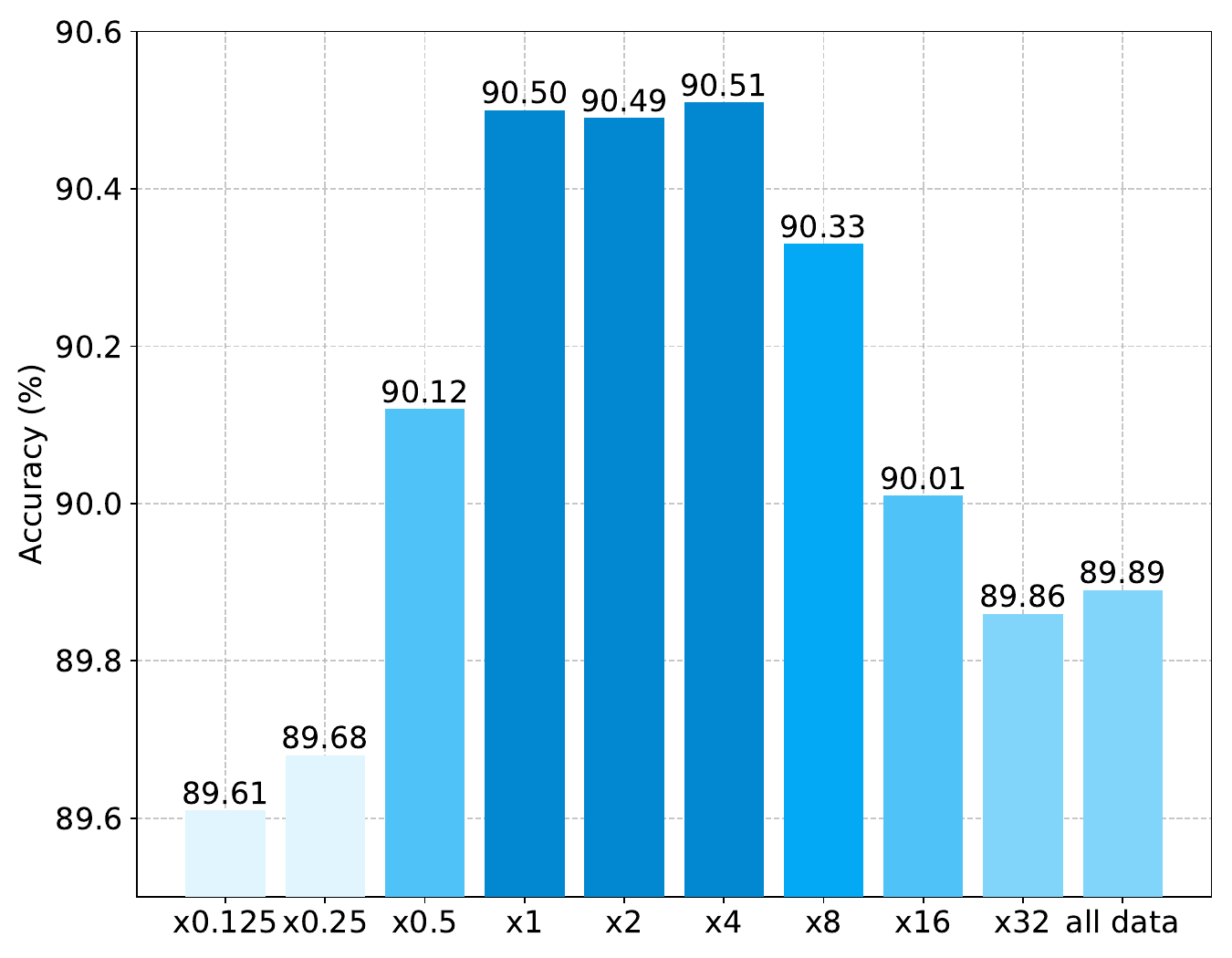}
    
  \caption{Impact of varying scales of sample size \( N_L \) relative to the average number of images per class on PACS dataset.} 
  \label{fig:datasize}
\end{minipage}\hfill
\begin{minipage}[t]{0.485\linewidth}
\centering
\includegraphics[width=1.0\textwidth]{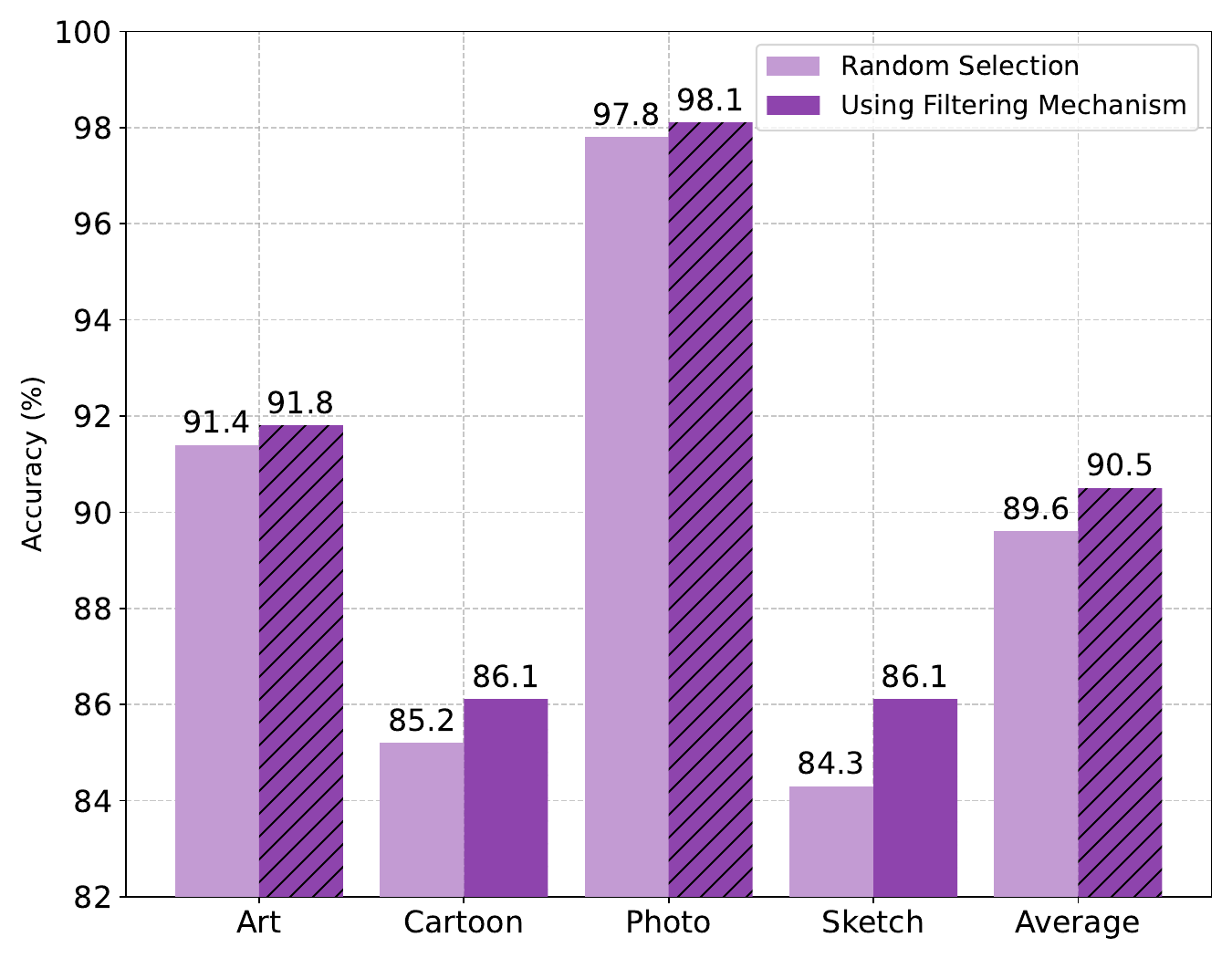}
% \caption{Accuracy (\%) Comparison: Random Selection vs. Proposed Filtering Strategy of FDS on accuracy (\%)}
\caption{Impact of using Random Selection vs. Proposed Filtering Strategy of FDS on PACS accuracy (\%).}
\label{fig:randomvsentropy}             
\end{minipage}
\vspace{-0.3cm}
\end{figure}

\begin{figure}[t]
  \centering
  \includegraphics[width=0.9\linewidth]{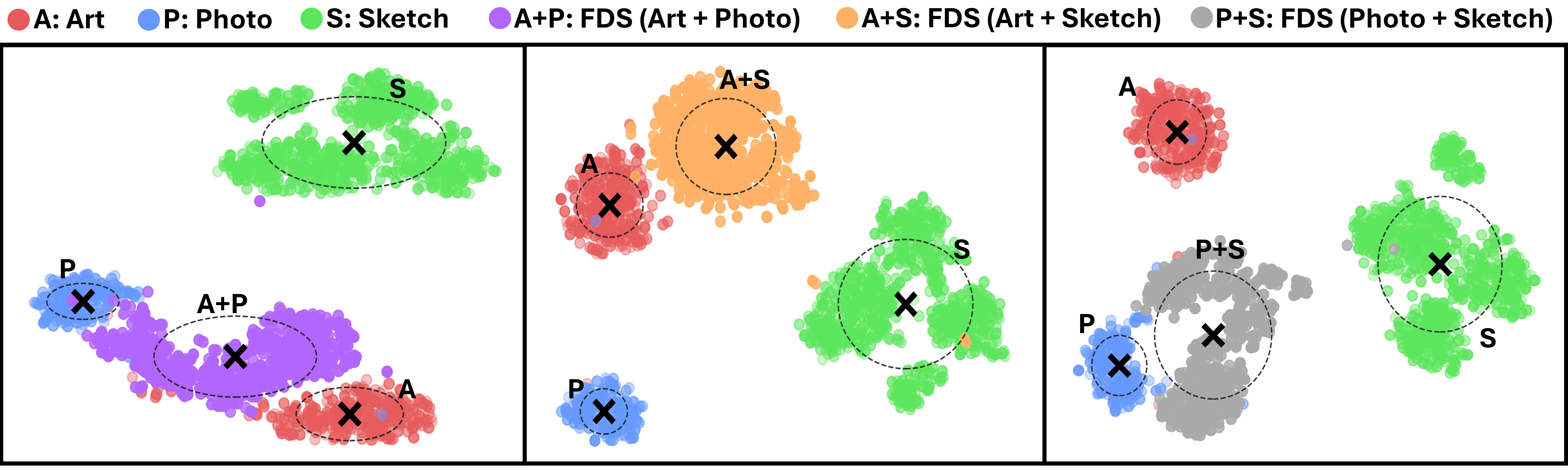}
  \caption{t-SNE plots showcasing the original ``giraffe'' class samples for the ``Art'', ``Photo'' and ``Sketch'' source domains of the PACS dataset, as well as the data generated with FDS.}
    \vspace{-0.3cm}
  \label{fig:diversity_tsne}
\end{figure}

\mypar{Impact of Sample Size.} 
Next, we assess the impact of the selected data size, denoted as $N_L$, on final performance. We consider the PACS dataset for this analysis, where the average number of images per class is 570. We systematically explore varying scales relative to this average class size, aiming to discern the optimal dataset size for enhancing OOD generalization. Our findings, detailed in Figure~\ref{fig:datasize}, reveal an initial improvement in OOD generalization with increased data size. However, excessively enlarging the dataset size begins to diminish the benefits of our filtering mechanism, as it incorporates a broader array of samples, including those that are overly simplistic and not conducive to model improvement.

\begin{table}[t]
    \vspace{-0.2cm}
    \centering
    \centering
    \tabcolsep=0.05cm
    \renewcommand{\arraystretch}{1} % Increase space between rows
    \adjustbox{max width=.85\linewidth}{
    \begin{tabular}{l|p{0.1cm}cp{0.1cm}cp{0.1cm}cp{0.1cm}|c}
    \toprule
     \textbf{Method}          && \textbf{PACS}   && \textbf{VLCS}             && \textbf{Office}          && \textbf{  ~Avg. }              \\
      \midrule
     Baseline (SWAD~\cite{cha2021swad})  && 88.11\ppm0.4 && 78.87\ppm0.5 && 70.34\ppm0.4   && 79.11  \\
     Filtered Based on Entropy           && 90.50\ppm0.4 && 79.33\ppm0.8 && 71.97\ppm0.3   && 80.60\\
     + Reject Semantic Loss     && \textbf{90.50\ppm0.3} && \textbf{79.73\ppm0.5} && \textbf{73.51\ppm0.5}   && \textbf{81.25}  \\ 
    \bottomrule
    \end{tabular}}
    \vspace{0.2cm}
    \caption{Impact of filtering strategy components of FDS on accuracy (\%), using three benchmarks.}
    \label{tab:onlycorrect}
\end{table}

\mypar{Filtering Mechanism.}
One might consider that %incorporating all generated samples could imbalance the augmented dataset compared to the original, suggesting that 
the enhanced accuracy observed with our filtering mechanism might be the result of constraining the sample size for a balanced final training set. To investigate this hypothesis, Figure~\ref{fig:randomvsentropy} contrasts the outcome of selecting $N_L$ samples at random from all generated images per class against employing our entropy-based filtering strategy. The consistent improvement in out-of-domain (OOD) generalization across all domains, facilitated by our filtering approach, underscores its effectiveness. 

Additionally, to examine the impact of excluding samples that deviate semantically which will be identified through misclassification by a classifier trained solely on the original dataset, we contrasted the accuracy between selections purely based on entropy and those refined this way in Table~\ref{tab:onlycorrect}. This comparison highlights the significance and efficiency of our filtering strategy in preserving semantic integrity.

\begin{table}[t]
    \vspace{-0.2cm}
    \centering
    \tabcolsep=0.2cm
    \adjustbox{max width=0.75\linewidth}{
      \begin{tabular}{lcccc|c}
        \toprule
        % & \multicolumn{5}{|c}{Accuracy\,(\%)} \\
        %Data Size (\#imgs/class) &  Art & Cartoon & Photo & Sketch & Avg.\\
        \multirow[b]{2}{*}{\bf Data} & \multicolumn{5}{c}{\bf Target Domains} \\
        \cmidrule(l{4pt}r{4pt}){2-6}
         &  \bf Art & \bf Cartoon & \bf Photo & \bf Sketch & \bf Avg.\\
        \midrule
        Original PACS & 0.39 & 0.87 & 0.54 & 0.99 & 0.70 \\
        Basic Gen. & 0.54 & 0.72 & 0.45 & 1.01 & 0.68 \\
        FDS & 0.44 & 0.58 & 0.49 & 0.92 & \textbf{0.61} \\
      \bottomrule
    \end{tabular}
}
\vspace{0.2cm}
\caption{Domain diversity metric~\cite{ye2022ood} between source domains and the target domain of the PACS dataset. ``\textbf{Basic Gen.}'' refers to generation without interpolation or filtering.}
\label{tab:domain_metric}
\end{table}

\mypar{Domain Diversity Visualization.}
To illustrate the effectiveness of our method, we present t-SNE plots of original PACS dataset samples and those generated by FDS in Figure~\ref{fig:diversity_tsne}. The t-SNE plots show distinct clusters for original source domains (here Art, Photo, Sketch) and highlight how the generated samples clearly bridge the gaps between these clusters, thus enhancing domain diversity. This expanded diversity is crucial for improving the generalization capabilities of models, as it ensures a broader spectrum of data distributions in the training set. By providing a continuous representation of the domain space, our method facilitates smoother transitions and better prepares models to handle unseen domains, ultimately contributing to more robust performance in real-world applications. Please check the supplementary materials for further details.

\begin{table}[!t]
    \centering
    \tabcolsep=0.2cm
    \adjustbox{max width=\linewidth}{
      \begin{tabular}{lcccc|c}
        \toprule
        % & \multicolumn{5}{|c}{Accuracy\,(\%)} \\
        %Data Size (\#imgs/class) &  Art & Cartoon & Photo & Sketch & Avg.\\
        \multirow[b]{2}{*}{\bf Method} & \multicolumn{5}{c}{\bf Source Domains} \\
        \cmidrule(l{4pt}r{4pt}){2-6}
         &  \bf C,\,S,\,P & \bf A,\,P,\,S & \bf A,\,C,\,S & \bf A,\,C,\,P & \bf Avg.\\
        \midrule
        ERM & 98.05\ppm0.5 & 96.06\ppm0.9 & 95.66\ppm0.3 & 96.83\ppm0.6 & 96.65 \\
        ERM + Dup. Aug.  & 98.01\ppm0.2 & 97.07\ppm0.3 & 96.46\ppm0.2 & 97.28\ppm0.3 & 97.21 \\
        ERM + Basic Gen. & 96.96\ppm0.4 & 96.74\ppm0.2 & 96.23\ppm0.2 & 97.37\ppm0.5 & 96.83 \\
        \rowcolor{gray!15}
        \textbf{ERM + FDS (ours)} & 97.77\ppm0.4 & 97.53\ppm0.5 & 96.71\ppm0.1 & 98.02\ppm0.4 & \textbf{97.51} \\
        \midrule
        SWAD & 98.48\ppm0.3 & 97.87\ppm0.5 & 97.61\ppm0.2 & 98.44\ppm0.0 & 98.10 \\
        SWAD + Dup. Aug. & 98.35\ppm0.2 & 97.86\ppm0.3 & 97.38\ppm0.1 & 98.16\ppm0.1 & 97.94 \\
        SWAD + Basic Gen. & 98.42\ppm0.4 & 98.23\ppm0.3 & 97.87\ppm0.5 & 98.64\ppm0.2 & 98.29 \\
        \rowcolor{gray!15}
        \textbf{SWAD + FDS (ours)} & 98.70\ppm0.3 & 98.25\ppm0.5 & 97.80\ppm0.4 & 98.61\ppm0.0 & \textbf{98.34} \\
        \bottomrule
      \end{tabular}
    }
\vspace{0.2cm}
\caption{Impact of FDS on in-domain PACS accuracy (\%). `A', `C', `P', and `S' refer to `Art', `Cartoon', `Photo', and `Sketch'.}
\label{tab:reg_updated}
\end{table}

\begin{table}[!t]
    \centering
    \tabcolsep=0.2cm
    \adjustbox{max width=\linewidth}{
      \begin{tabular}{lcccc|c}
        \toprule
        % & \multicolumn{5}{|c}{Accuracy\,(\%)} \\
        %Data Size (\#imgs/class) &  Art & Cartoon & Photo & Sketch & Avg.\\
        \multirow[b]{2}{*}{\bf Method} & \multicolumn{5}{c}{\bf Target Domains} \\
        \cmidrule(l{4pt}r{4pt}){2-6}
         &  \bf Art & \bf Cartoon & \bf Photo & \bf Sketch & \bf Avg.\\
        \midrule
        ERM & 86.94\ppm0.6 & 80.21\ppm0.7 & 96.61\ppm0.4 & 74.45\ppm2.9 & 84.30 \\
        ERM + Dup. Aug. & 85.34\ppm0.9 & 80.99\ppm0.9 & 94.98\ppm0.8 & 76.88\ppm2.0 & 84.55 \\
        ERM + Basic Gen. & 87.21\ppm0.3 & 80.90\ppm1.9 & 95.71\ppm0.4 & 80.31\ppm2.3 & 86.03 \\
        \rowcolor{gray!15}
        \textbf{ERM + FDS (ours)} & 90.69\ppm0.9 & 84.19\ppm0.6 & 97.21\ppm0.1 & 82.99\ppm0.4 & \textbf{88.77} \\
        \midrule
        SWAD & 89.49\ppm0.2 & 83.65\ppm0.4 & 97.25\ppm0.2 & 82.06\ppm1.0 & 88.11 \\
        SWAD + Dup. Aug. & 89.55\ppm0.3 & 83.00\ppm1.7 & 97.65\ppm0.2 & 81.86\ppm1.1 & 88.02 \\
        SWAD + Basic Gen. & 89.87\ppm0.1 & 85.59\ppm0.6 & 97.50\ppm0.3 & 83.07\ppm0.4 & 89.01 \\
        \rowcolor{gray!15}
        \textbf{SWAD + FDS (ours)} & 91.80\ppm0.3 & 86.03\ppm0.8 & 98.05\ppm0.2 & 86.11\ppm0.1 & \textbf{90.50} \\
        \bottomrule
      \end{tabular}
    }
    \vspace{0.2cm}
    \caption{Leave-one-out accuracy (\%) of FDS compared to other augmentation strategies.}
    \label{tab:reg_out_domain}

\end{table}

\mypar{Domain Diversity Quantification.}
We further validate the effectiveness of our method using a domain diversity metric based on the methodology proposed in~\cite{ye2022ood}. This metric quantifies the diversity shift between the source domains and the target domain, providing valuable insight into how the newly generated domains using our method can improve generalization on unseen domains. Table~\ref{tab:domain_metric} presents the domain diversity metric for the original PACS dataset, comparing it with samples generated using the diffusion model in its basic form (no interpolation and no filter) and our FDS method (including interpolation and filtering). We observe that our FDS method results in a lower diversity metric compared to using the original PACS dataset or the samples generated with basic generation, suggesting that our method more effectively promotes generalization to new, unseen domains.

\mypar{In-Domain Regularization Effect.} In this section, we study the impact of incorporating images generated by our method on the network's accuracy, when testing on the same domains as those used for training. To provide a more comprehensive analysis, we adopted an 80/10/10 split for the source domains\footnote{\ This setting was only used for in-domain experiments, while the rest of the paper follows the standard 80/20 setting as used in DomainBed.}. We selected the best model on the validation set and reported its performance on the held-out in-domain test set. As indicated in Table~\ref{tab:reg_updated}, our FDS approach surpasses the baseline accuracy in this setup for both the ERM and SWAD backbones. Additionally, we compared our method to two alternative strategies: \emph{i}) duplicating and augmenting original samples (Dup. Aug.), and \emph{ii}) generating synthetic in-domain images (same number of images as FDS)  without domain interpolation (Basic Gen.). While these strategies also improve the performance, FDS achieves the highest accuracy due to its ability to generate more diverse and challenging pseudo-domains, rather than simply increasing dataset size. This confirms the value of our method not only in OOD conditions but also in standard in-domain validation, aligning with the principles of Vicinal Risk Minimization (VRM)~\cite{chapelle2000vicinal}. Hence, our method may also be viewed as a regularization strategy, suitable for a broad spectrum of applications. For completeness, we also report the out-domain performance of FDS and these strategies in Table \ref{tab:reg_out_domain}. As can be seen, the beneficial impact of FDS on OOD generalization is not simply the result of increasing the dataset size.

\section{Conclusion}
\label{sec:conclusion}
This work presented FDS, a domain generalization (DG) technique that leverages diffusion models for domain mixing, generating a diverse set of images to bridge the domain gap between source domains distribution. We also proposed an entropy-based filtering strategy that enriches the pseudo-novel generated set with images that test the limits of classifiers trained on original data, thereby boosting generalization. Our extensive experiments across multiple benchmarks demonstrate that our method not only surpasses existing DG techniques but also sets new records for accuracy.
Our analysis indicate that our approach contributes to more stable training processes when confronted with domain shifts and serves effectively as a regularization method in in-domain contexts. Notably, our technique consistently enhances performance across diverse scenarios, from realistic photos to sketches. While we exploited our trained diffusion model for covering the domain gap, more sophisticated techniques could be considered. For instance, future work could investigate the idea of generating pseudo-novel distributions via extrapolation in the domain space, in addition to interpolation.

\newpage

\section*{\centering FDS: Feedback-guided Domain Synthesis with Multi-Source Conditional Diffusion Models for Domain Generalization -- Supplementary Material}

%%%%%%%%%%%%%%%%%%%%%%%%%%%%%%%%%%%%
\appendix

\vspace{1cm}
\section{Implementation}
Our proposed FDS method is built using the Python language and the PyTorch framework. We utlized four NVIDIA A100 GPUs for all our experiments. For initializing our models, we utilize the original Stable Diffusion version 1.5 as our initial weight ~\cite{rombach2022high}. The key hyperparameter configurations employed for training these diffusion models and generating new domains are detailed in Tables~\ref{tab:hyperparameters_dm} and~\ref{tab:hyperparameters_gen}, respectively.

Furthermore, for classifier training, we adhere to the methodologies and parameter settings described by Cha et al.~\cite{cha2021swad}, ensuring consistency and reproducibility in our experimental setup. The original implementation and instructions for reproducing our results are accessible via \url{https://github.com/Mehrdad-Noori/FDS}.

\section{Additional Ablation}

\noindent \textbf{Selection/Filtering.} In this section, we provide visual examples to show the efficacy of our synthetic sample selection and filtering mechanism. As mentioned in the method section, this mechanism is intricately designed to scrutinize the generated images through two lenses: the alignment of the predicted class with the intended label, and the entropy indicating the prediction's uncertainty.

\begin{figure*}[t]
  \centering
  \includegraphics[width=1\textwidth]{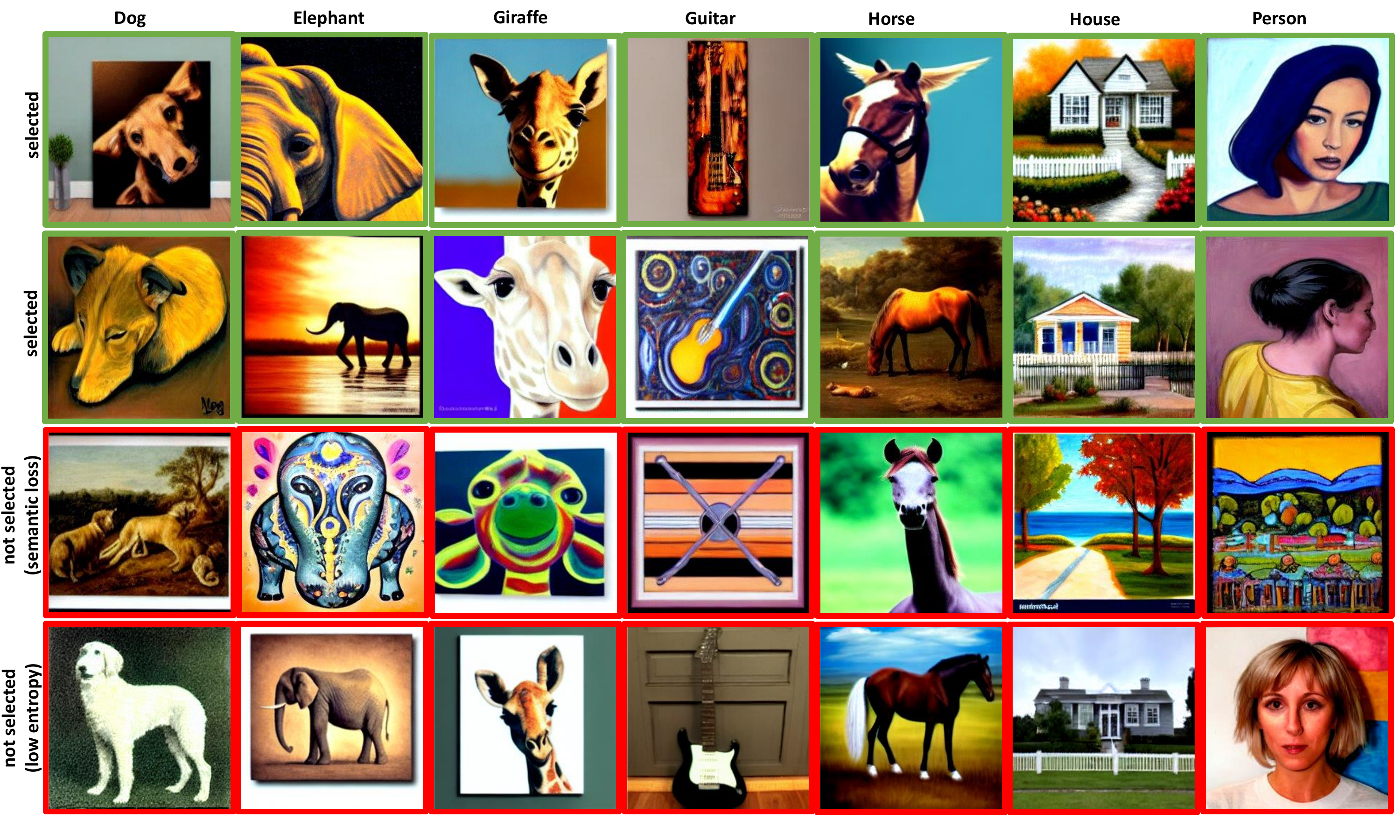}
  \caption{Synthetic images from interpolating between \emph{``art''} and \emph{``photo''} domains of PACS, with \textcolor{green}{selected} images showcasing a blend of artistic and realistic features (top two rows) and \textcolor{red}{non-selected} images (bottom rows) due to class mismatches and low entropy.}
  \label{fig:s_ap}
\end{figure*}

\begin{figure*}[t]
  \centering
  \includegraphics[width=1\textwidth]{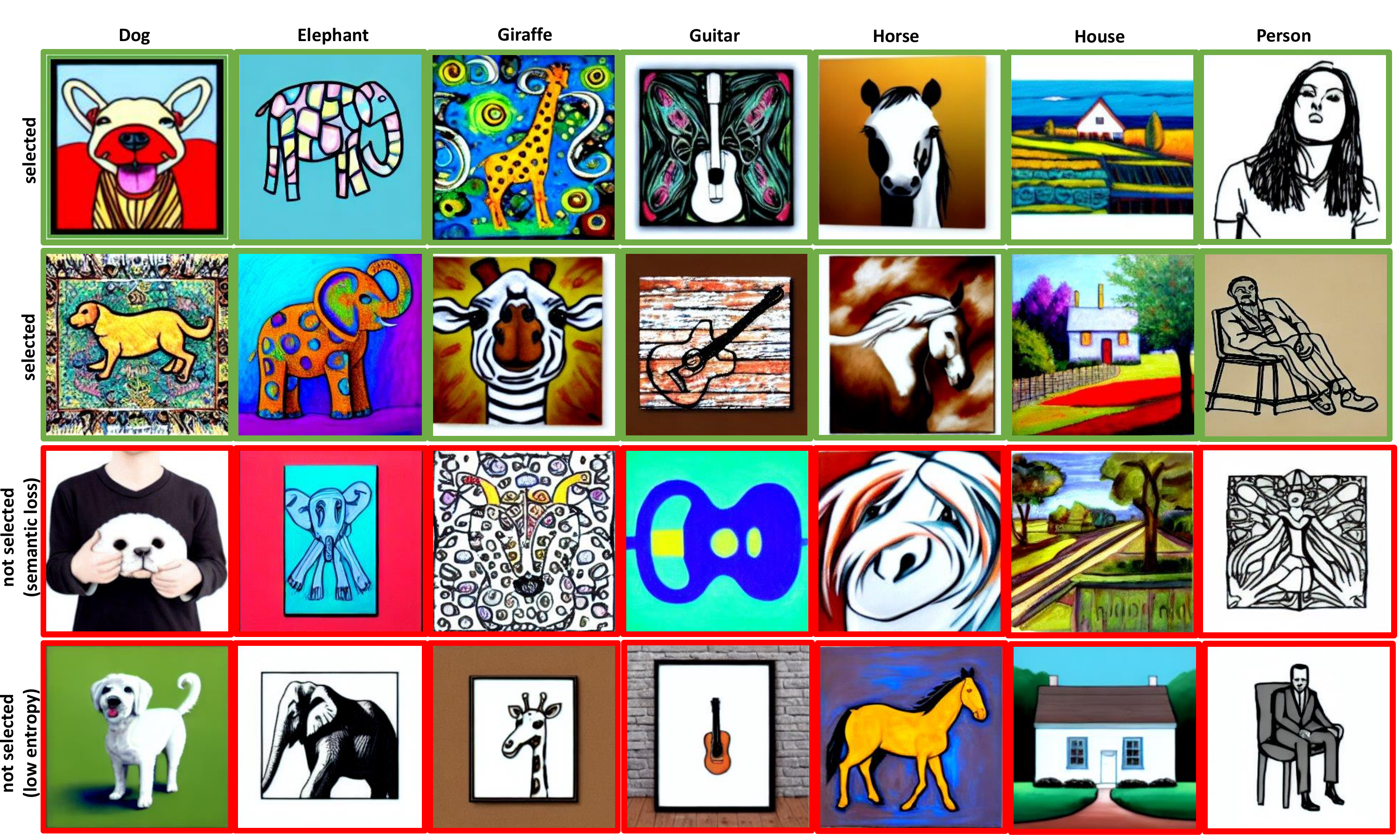}
  \caption{Interpolation between \emph{``art''} and \emph{``sketch''} in PACS highlights \textcolor{green}{selected} images (top rows) merging textures and outlines, and \textcolor{red}{non-selected} images (bottom rows) for failing selection criteria.}
  \label{fig:s_as}
\end{figure*}

\begin{figure*}[t]
  \centering
  \includegraphics[width=1\textwidth]{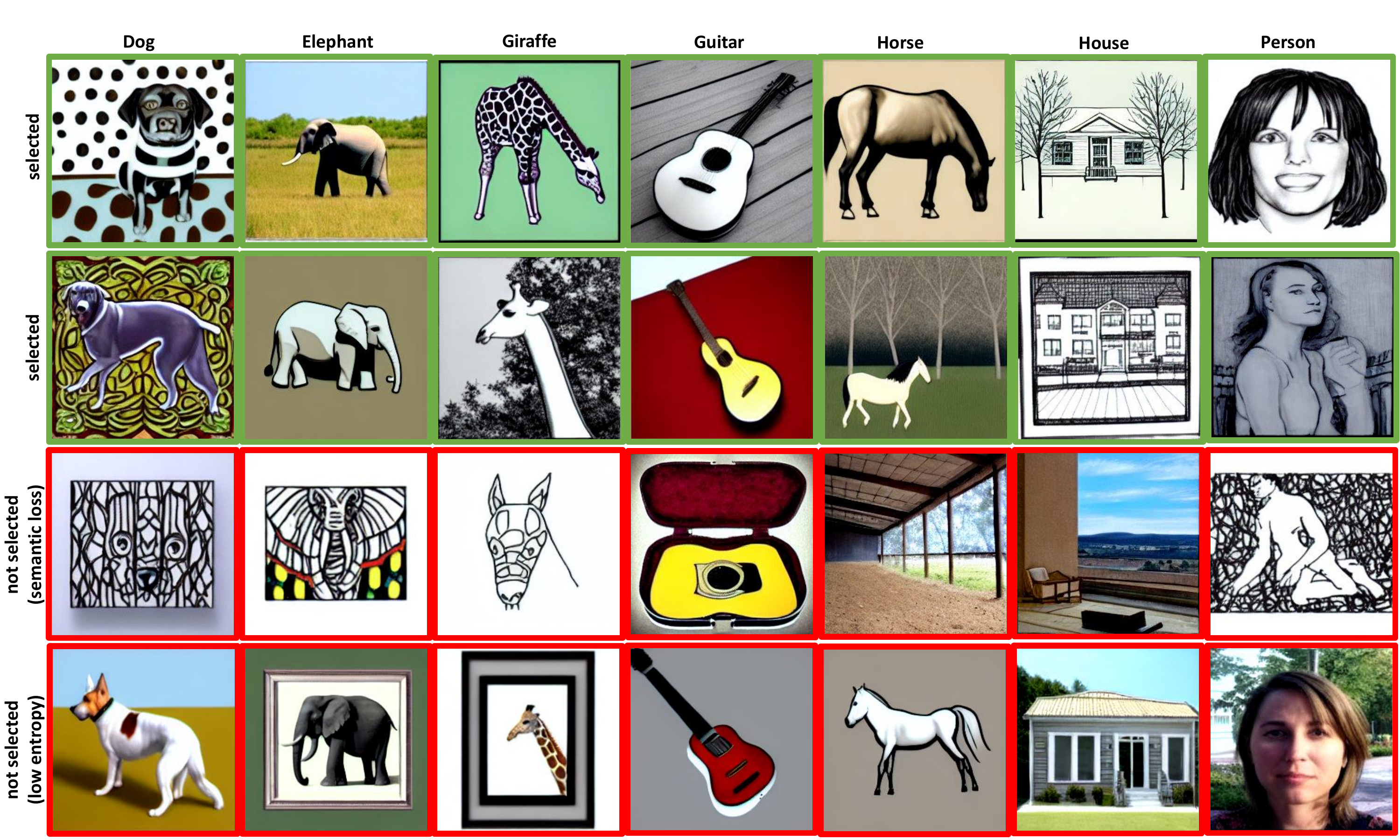}
  \caption{Results from \emph{``photo''} and \emph{``sketch''} domain interpolation in PACS, with \textcolor{green}{selected} synthetic images (top rows) and \textcolor{red}{non-selected} due to predictability and class misalignment (bottom rows).}
  \label{fig:s_ps}
\end{figure*}

The Figures~\ref{fig:s_ap},~\ref{fig:s_as},~\ref{fig:s_ps} showcase a set of images generated from interpolations between two domains. Specifically, the diffusion model is trained on \emph{``art''}, \emph{``sketch''}, and \emph{``photo''} of the PACS dataset, and the selected images, demonstrated in the first two rows, exemplify successful blends of domain characteristics, embodying a balanced mixture that enriches the training data with novel, domain-bridging examples. These images were chosen based on their ability to meet our criteria: correct class prediction aligned with high entropy scores. The third and fourth rows highlight the filtering aspect of our mechanism, displaying images not selected due to class mismatches and low entropy, respectively. This visual demonstration underlines the pivotal role of our selection/filtering process in refining the synthetic dataset, ensuring only the most challenging and domain-representative samples are utilized for model training. Through this approach, we aim to significantly bolster the model's capacity to generalize across diverse visual domains.

\begin{figure*}[t]
  \centering
  \includegraphics[width=0.95\textwidth]{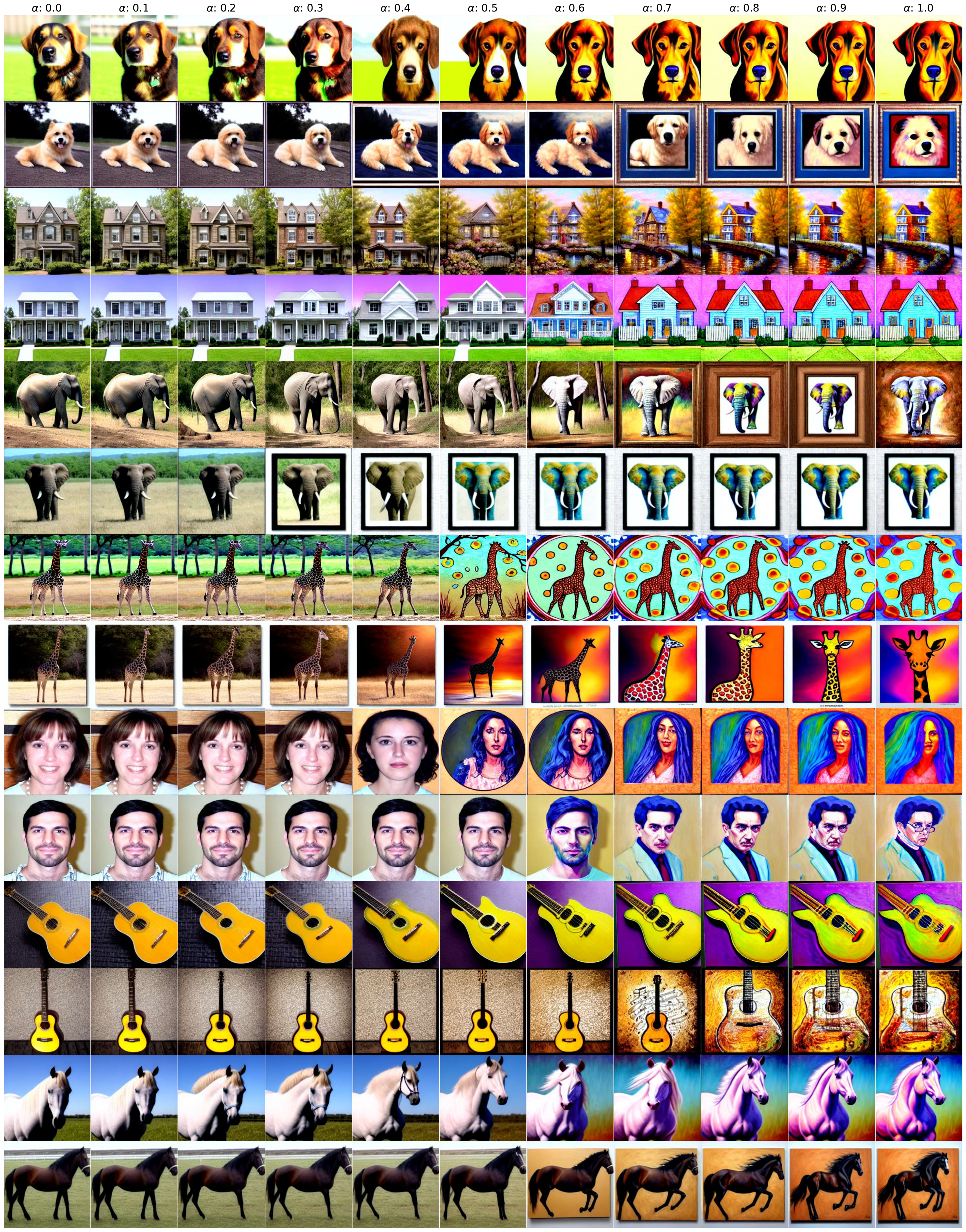}
  \caption{Inter-domain Transition from \emph{``photo''} to \emph{``art''}. This sequence illustrates how varying \(\alpha\) from 0.0 (purely photorealistic images) to 1.0 (purely artistic representations) enables the model to seamlessly blend photographic realism with artistic expression, demonstrating a smooth progression from real-world imagery to stylized art.}
  \label{fig:t_ap}
\end{figure*}

\begin{figure*}[t]
  \centering
  \includegraphics[width=0.95\textwidth]{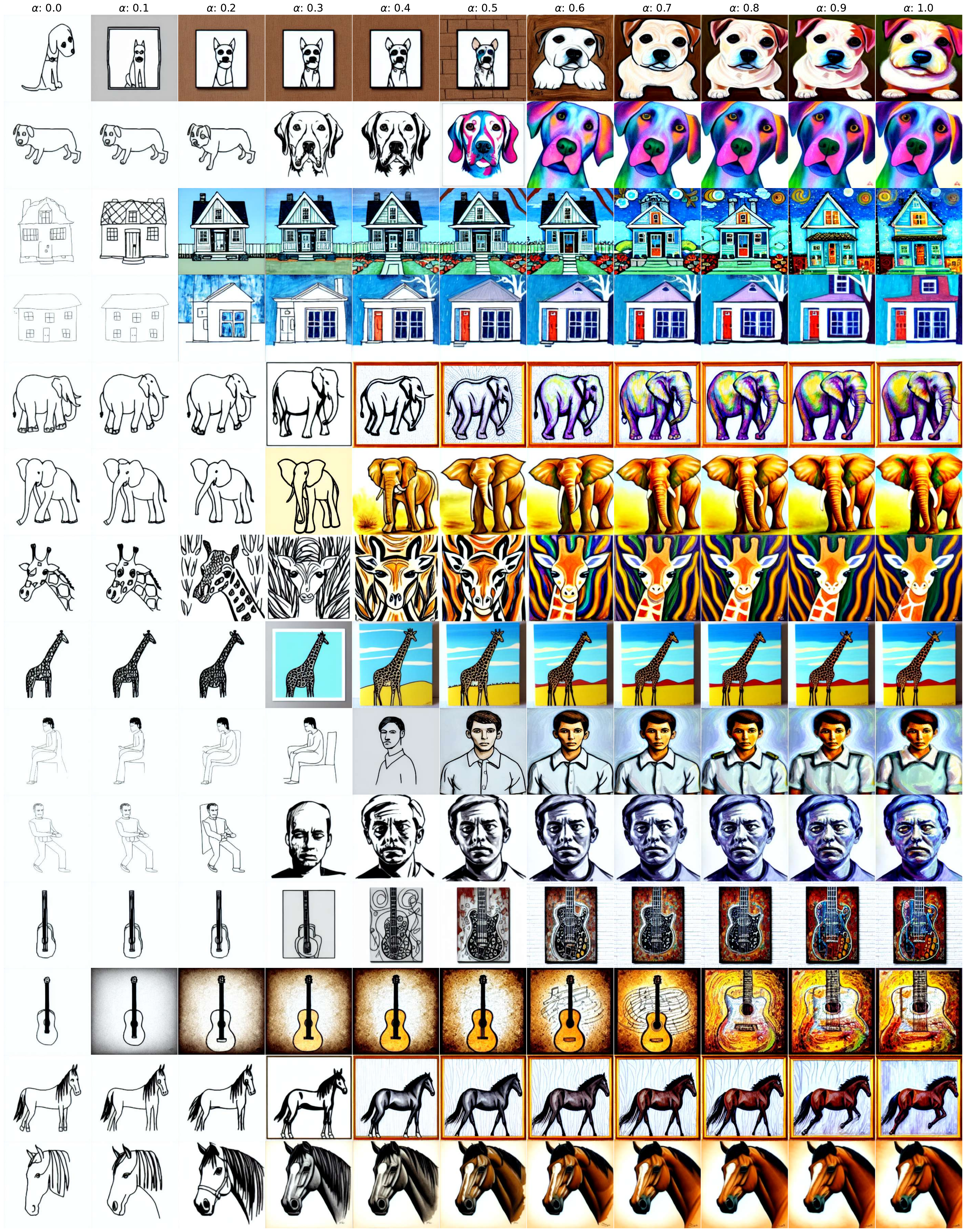}
  \caption{Inter-domain Transition from \emph{``sketch''} to \emph{``art''}. Displayed here is the transformation that occurs as \(\alpha\) is adjusted, beginning with 0.0 (pure sketches) and moving towards 1.0 (fully art-inspired images). The model effectively infuses basic sketches with complex textures and colors, transitioning from minimalistic line art to detailed and vibrant artistic images.}
  \label{fig:t_as}
\end{figure*}

\begin{figure*}[t]
  \centering
  \includegraphics[width=0.95\textwidth]{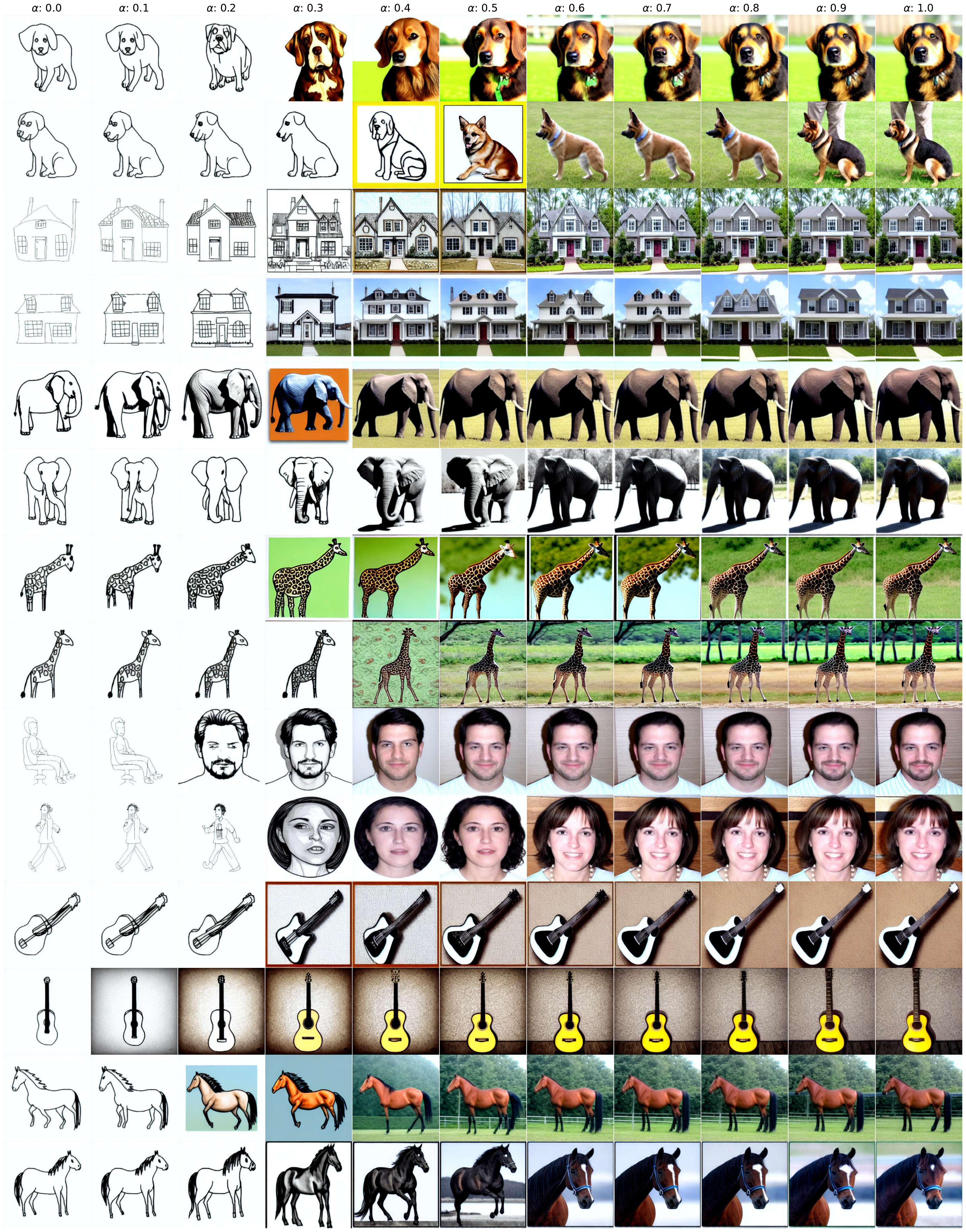}
  \caption{Inter-domain Transition from \emph{``sketch''} to \emph{``photo''}. This figure demonstrates the capability of the model to morph sketches into photorealistic images by altering \(\alpha\) from 0.0 (entirely sketch-based) to 1.0 (completely photorealistic). The transition highlights the model's proficiency in enriching simple outlines with lifelike details and textures, bridging the gap between abstract sketches and reality.}
  \label{fig:t_ps}
\end{figure*}

\noindent \textbf{Inter-domain Transition.} In this section, we demonstrate the model's ability to navigate between distinct visual domains, a capability enabled by adjusting the mix coefficient \(\alpha\). Trained on multiple source domains, our model can generate images that blend the unique attributes of each source domain. By varying \(\alpha\) from 0.0 to 1.0, we enable smooth transitions between two source domains, where \(\alpha=0.0\) and \(\alpha=1.0\) correspond to generating pure images of the first and second domain, respectively. As an example, we illustrated this ability for our model trained on the PACS sources' \emph{``art''}, \emph{``sketch''}, and \emph{``photo''}.
These domain transitions are illustrated in the figures, showcasing transitions from \emph{``photo''} to \emph{``art''} domain in Figure~\ref{fig:t_ap}, \emph{``sketch''} to \emph{``art''} domain in Figure~\ref{fig:t_as}, and \emph{``sketch''} to \emph{``photo''} domain in Figure~\ref{fig:t_ps}, respectively. The examples provided highlight the effectiveness of our interpolation method in producing images that incorporate the distinctive features of the mixed domains, thus affirming the model's capability to generate novel and coherent visual content that bridges the attributes of its training domains. Note that in all of our generation experiments, we constrained \(\alpha\) to the range of 0.3 to 0.7 to ensure the generated images optimally embody the characteristics of the two mixing domains, as detailed in Table~\ref{tab:hyperparameters_gen}.

\begin{table}[t!]
\centering

\small
\renewcommand{\arraystretch}{1.0} % Adjust the space between the table rows
\setlength{\tabcolsep}{1.7pt}
\begin{tabular}{lc}
\toprule
Config & Value \\ 
\midrule
Number of GPUs              & 4               \\
Learning rate               & 1e-4            \\
Learning rate scheduler     & LambdaLinear    \\
Batch size                  & 96 (24 per GPU) \\
Precision                   & FP16            \\
Max training steps          & 10000            \\
Denosing timesteps          & 1000            \\
Sampler                     & DDPM~\cite{ho2020denoising}            \\
Autoencoder input size      & 256 x 256 x 3   \\
Latent diffusion input size & 32 x 32 x 4     \\ 
\bottomrule
\end{tabular}
\vspace{0.2cm}
\caption{Hyperparameter Configuration for Training Diffusion Models.}
\label{tab:hyperparameters_dm}
\end{table}

\begin{table}[t!]
\centering
\small
\renewcommand{\arraystretch}{1.0} % Adjust the space between the table rows
\setlength{\tabcolsep}{1.7pt}
\begin{tabular}{lc}
\toprule
Config & Value \\ 
\midrule
Sampler                                    & DDIM~\cite{song2020denoising}                                 \\
Denosing timesteps                         & 50                                   \\
Classifier-free guidance (CFG)             & Randomly from {[}5, 6{]}    \\
Mix coefficient $\alpha$                   & Randomly from {[}0.3, 0.7{]} \\
Mix timestep $T$                           & Randomly from {[}20, 45{]}  \\
Generated images (PACS)                    & 32k per class                        \\
Generated images (VLCS)                    & 32k per class                        \\
Generated images (OfficeHome)              & 16k per class                        \\
\bottomrule
\end{tabular}
\vspace{0.2cm}
\caption{Hyperparameter Configuration for Generating New Domains.}
\label{tab:hyperparameters_gen}
\end{table}

\noindent \textbf{Number of Generated Domains}
The impact of varying the number of generated domains on model performance was rigorously evaluated, as summarized in Table~\ref{tab:num_domains}. This analysis aimed to understand how different combinations of augmented domains influence the overall accuracy across various dataset domains such as Art, Cartoon, Photo, and Sketch. By integrating diverse domain combinations, identified by IDs (as defined in Table~\ref{tab:ID_definitions}), we observed improvement gain when we add more generated domain of different combinations. Notably, all possible combinations of augmented domains (3 new domains for PACS, VLCS and OfficeHome) were utilized as the final method, leveraging the full spectrum of available data domains.

\begin{table*}[t]

    \centering
    \tabcolsep=0.2cm
    \adjustbox{max width=.8\textwidth}{
      \begin{tabular}{l>{\centering\arraybackslash}p{2.5cm}cccccc}
        \toprule
        \multirow{2}[4]{*}{\bf Method} & \multirow{2}[4]{2.5cm}{\centering\bf Augmented Domains} & \multicolumn{5}{c}{\bf Accuracy (\%)} \\
        \cmidrule(lr){3-7}
         & & \bf Art & \bf Cartoon & \bf Photo & \bf Sketch & \bf Avg. \\
        \midrule
        SWAD (reproduced)   & --- & 89.49\ppm0.2 & 83.65\ppm0.4 & 97.25\ppm0.2 & 82.06\ppm1.0 & 88.11\ppm0.45 \\
        SWAD + FDS  & ID0 & 91.03\ppm0.5 & 83.87\ppm0.6 & 97.75\ppm0.3 & 85.77\ppm0.4 & 89.61\ppm0.30 \\
        SWAD + FDS  & ID1 & 91.01\ppm0.6 & 85.06\ppm1.3 & 97.90\ppm0.3 & 83.64\ppm0.4 & 89.40\ppm0.65 \\
        SWAD + FDS  & ID2 & 91.46\ppm0.3 & 85.22\ppm0.8 & 97.88\ppm0.2 & 84.27\ppm0.3 & 89.71\ppm0.40 \\
        SWAD + FDS  & ID0 + ID1 & 91.52\ppm0.0 & 85.87\ppm0.7 & 98.03\ppm0.3 & 85.70\ppm1.0 & 90.28\ppm0.50 \\
        SWAD + FDS  & ID1 + ID2 & 91.62\ppm0.8 & 85.57\ppm0.4 & 98.20\ppm0.3 & 83.88\ppm0.6 & 89.82\ppm0.53 \\
        SWAD + FDS  & ID0 + ID2 & 91.52\ppm0.1 & 84.54\ppm0.5 & \textbf{98.28\ppm0.1} & \textbf{86.45\ppm0.8} & 90.20\ppm0.38 \\ 
        SWAD + FDS  & ID0 + ID1 + ID2 & \textbf{91.80\ppm0.3} & \textbf{86.03\ppm0.8} & 98.05\ppm0.2 & 86.11\ppm0.1 & \textbf{90.50\ppm0.35} \\
      \bottomrule
    \end{tabular}
    }
    \caption{Analysis of the impact of utilizing different numbers/combinations of generated domains on final model performance across the PACS dataset domains (Leave-one-out accuracy). For definitions of each augmented domain (ID0, ID1, ID2), see Table~\ref{tab:ID_definitions}.}
    \label{tab:num_domains}
\end{table*}

\begin{table*}[t]

    \centering
    \tabcolsep=0.11cm
    \adjustbox{max width=\linewidth}{
        \begin{tabular}{>{\centering\arraybackslash}m{2cm}cccc}
            \toprule
            \bf Augmented Domains & \bf Art & \bf Cartoon & \bf Photo & \bf Sketch \\
            \midrule
            ID0 & Cartoon + Photo & Art + Photo & Art + Cartoon & Art + Cartoon \\
            ID1 & Cartoon + Sketch & Art + Sketch & Art + Sketch & Art + Photo \\
            ID2 & Photo + Sketch & Photo + Sketch & Cartoon + Sketch & Cartoon + Photo \\
            \bottomrule
        \end{tabular}
    }
    \caption{Explanation of augmented domains ID definitions for each target domain of PACS dataset.}
    \label{tab:ID_definitions}
\end{table*}

\begin{figure}[!t]
  \centering
  \includegraphics[width=0.9\linewidth]{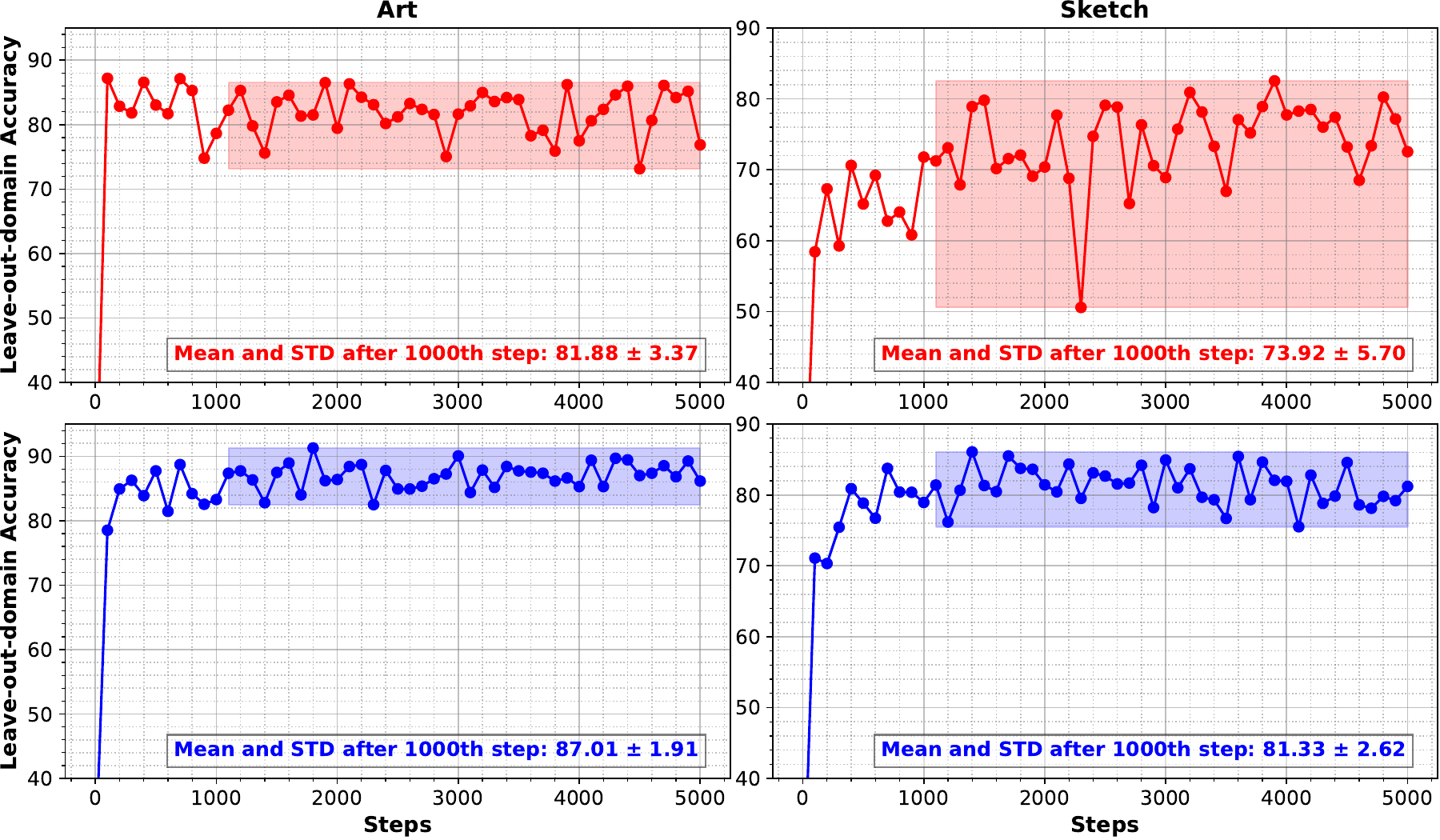}
  \caption{Accuracy (\%) across training steps: Comparison between ERM (top row) vs. FDS (bottom row) in "Art" and "Sketch" domains of PACS dataset.}
  \vspace{-0.3cm}
  \label{fig:stable}
\end{figure}

\mypar{Stability Analysis.} 
In this section, we demonstrate the performance of our model across different stages of training within two domains of the PACS dataset, depicted in Figure~\ref{fig:stable}. It is important to note that these test accuracies \emph{were not used} in the selection of the best-performing model mentioned in earlier sections and all of our experiments follow leave-one-out settings suggested by DomainBed. The results indicate that our model achieves higher stability and better mean accuracy with lower standard deviation compared to the ERM trained on original data. Note that we cannot plot the figures for SWAD since it is a WA of ERM and does not have individual training curves. These results demonstrate the robustness and stability of our model during training, which is crucial for domain generalization algorithms.

\begin{figure*}[t]
  \centering
  \hspace{1cm}
  \includegraphics[width=0.85\textwidth]{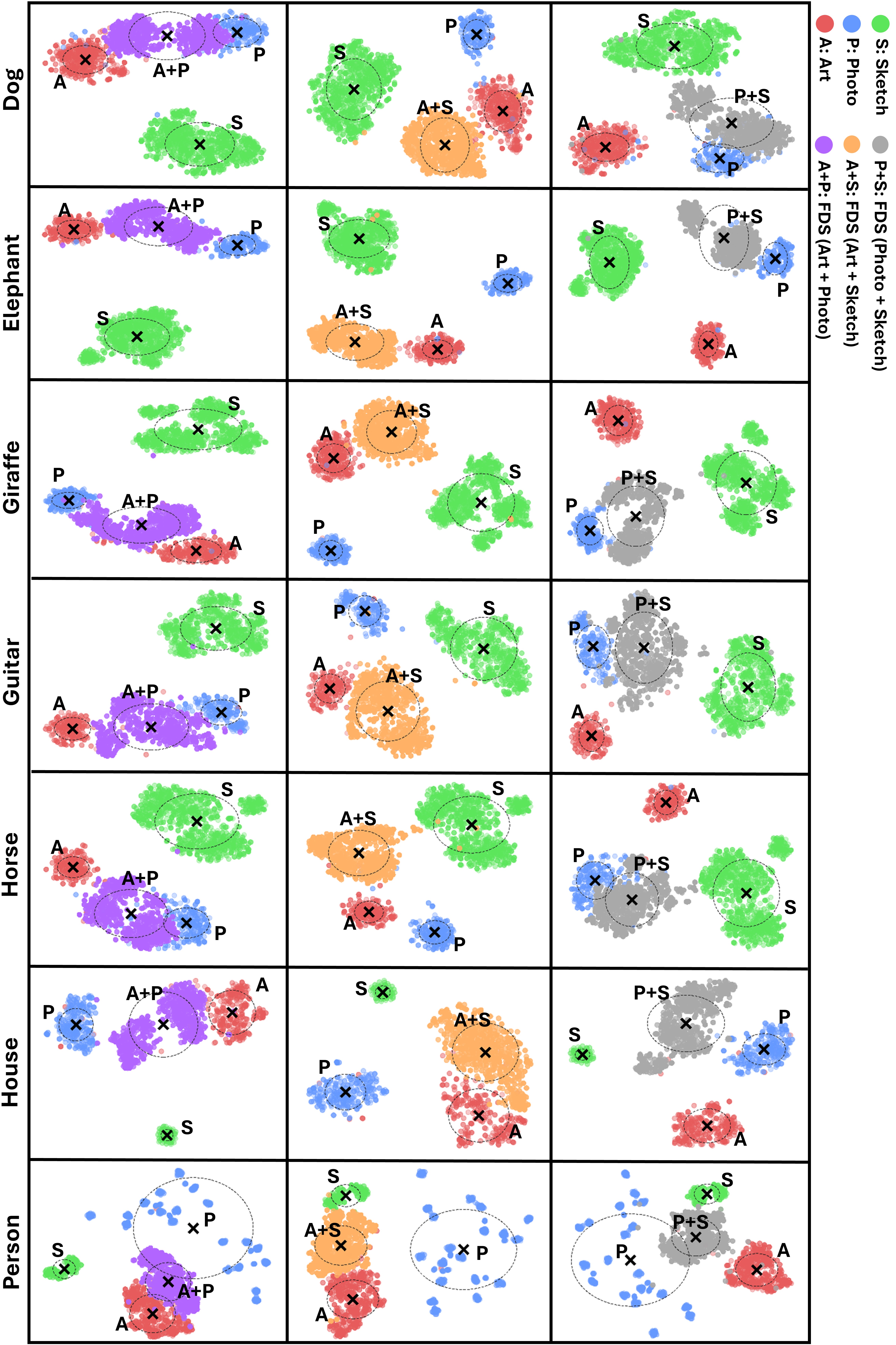}
  \caption{t-SNE visualization of all classes from the PACS dataset, showing the distribution of original source domains (Art, Photo, Sketch) and FDS ones.}
  \label{fig:tsne_fds}
\end{figure*}

\mypar{t-SNE Visualizations.}
This section presents a comprehensive t-SNE analysis for all classes in the PACS dataset, demonstrating the effectiveness of the FDS method in generating diverse, high-quality samples. The plots are provided in Figure~\ref{fig:tsne_fds}. Each t-SNE plot illustrates the distribution of both original and FDS-generated samples across different domains. The results shown here are based on our diffusion model trained on the ``Art,'' ``Photo,'' and ``Sketch'' source domains from the PACS dataset.
To create these visualizations, we extracted features using the CLIP vision encoder~\cite{radford2021learning}. Each class in the PACS dataset is represented as distinct clusters, with ``x'' markers indicating the location of the average representation of each domains. These averages serve as a reference to assess how well the FDS-generated samples are compared with the original domains.
These plots demonstrate how FDS enables smooth transitions between domains by interpolating between domain characteristics. This ability to generate synthetic data across a broad spectrum of domain representations improves the diversity of training data and enhances the model's generalization ability. By covering a wider range of the domain space, FDS helps the model better handle unseen domains, making it more robust in real-world applications. These visualizations also suggest that the generated domains can be viewed as new pseudo-domains, as the FDS samples exhibit distributions distinct from their original sources domains. This additional diversity is critical for training models capable of generalizing beyond the source domains.

\begin{figure*}[t]
  \centering
  \includegraphics[width=0.85\textwidth]{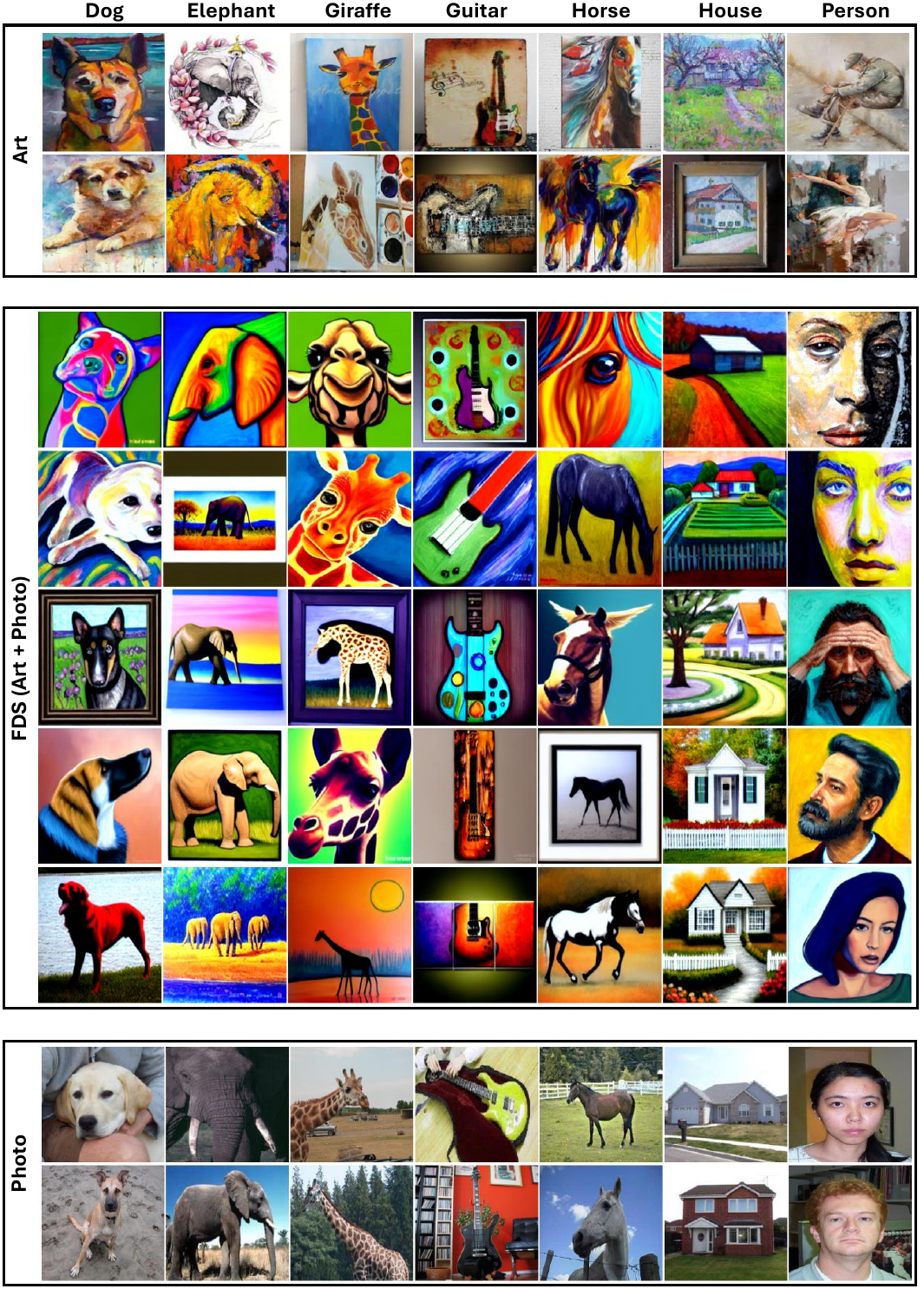}
  \caption{Visual comparison of original "Art" and "Photo" samples from PACS with synthetic images generated by FDS (Art + Photo). The middle section illustrates how FDS combines visual elements from both domains, producing diverse, domain-bridging images.}
  \label{fig:fds_ap}
\end{figure*}

\begin{figure*}[t]
  \centering
  \includegraphics[width=0.85\textwidth]{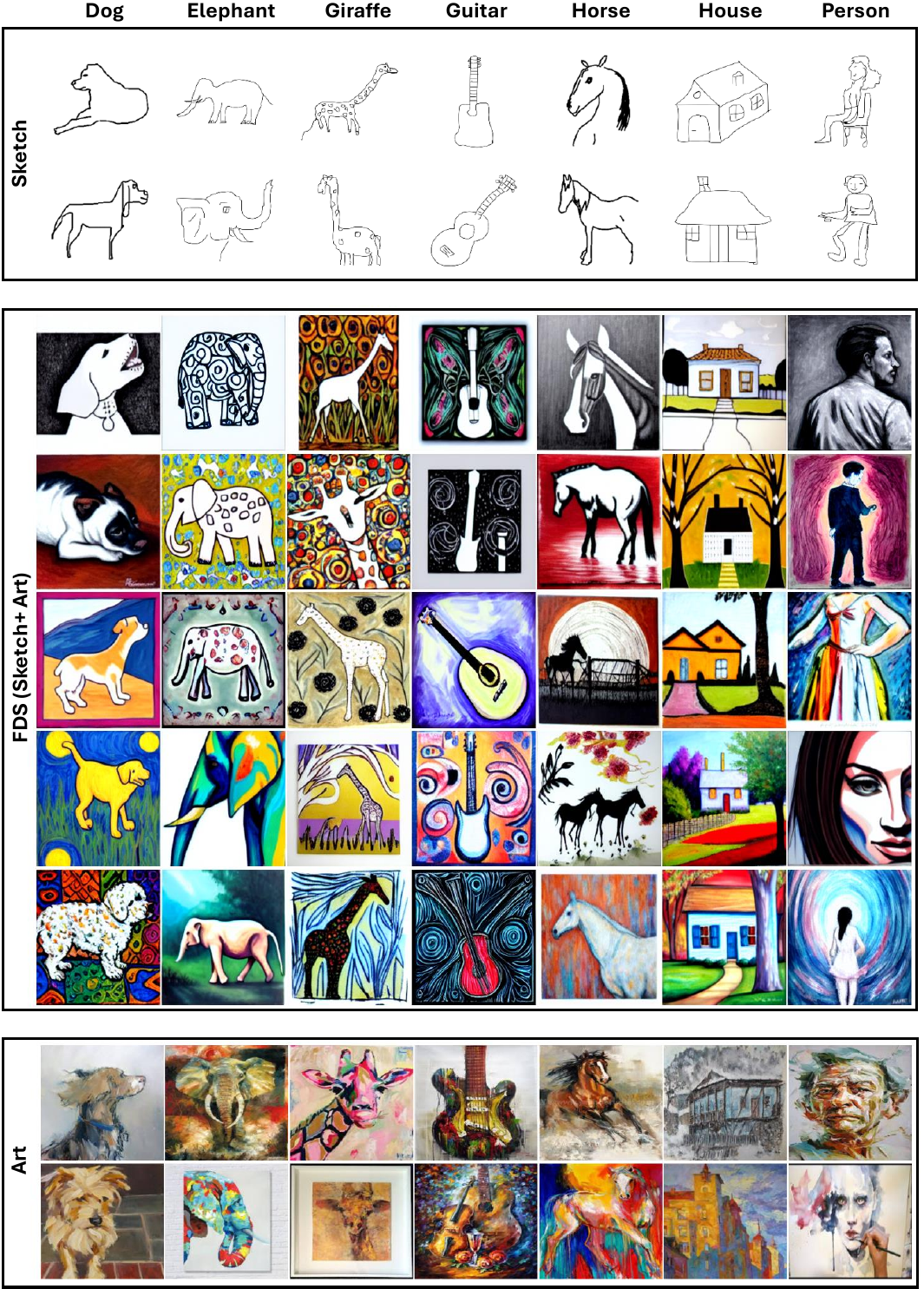}
  \caption{Visual comparison of original "Sketch" and "Art" samples from PACS with synthetic images generated by FDS (Sketch + Art). The generated images in the middle section showcase a blend of artistic textures and sketched outlines.}
  \label{fig:fds_sa}
\end{figure*}

\begin{figure*}[t]
  \centering
  \includegraphics[width=0.85\textwidth]{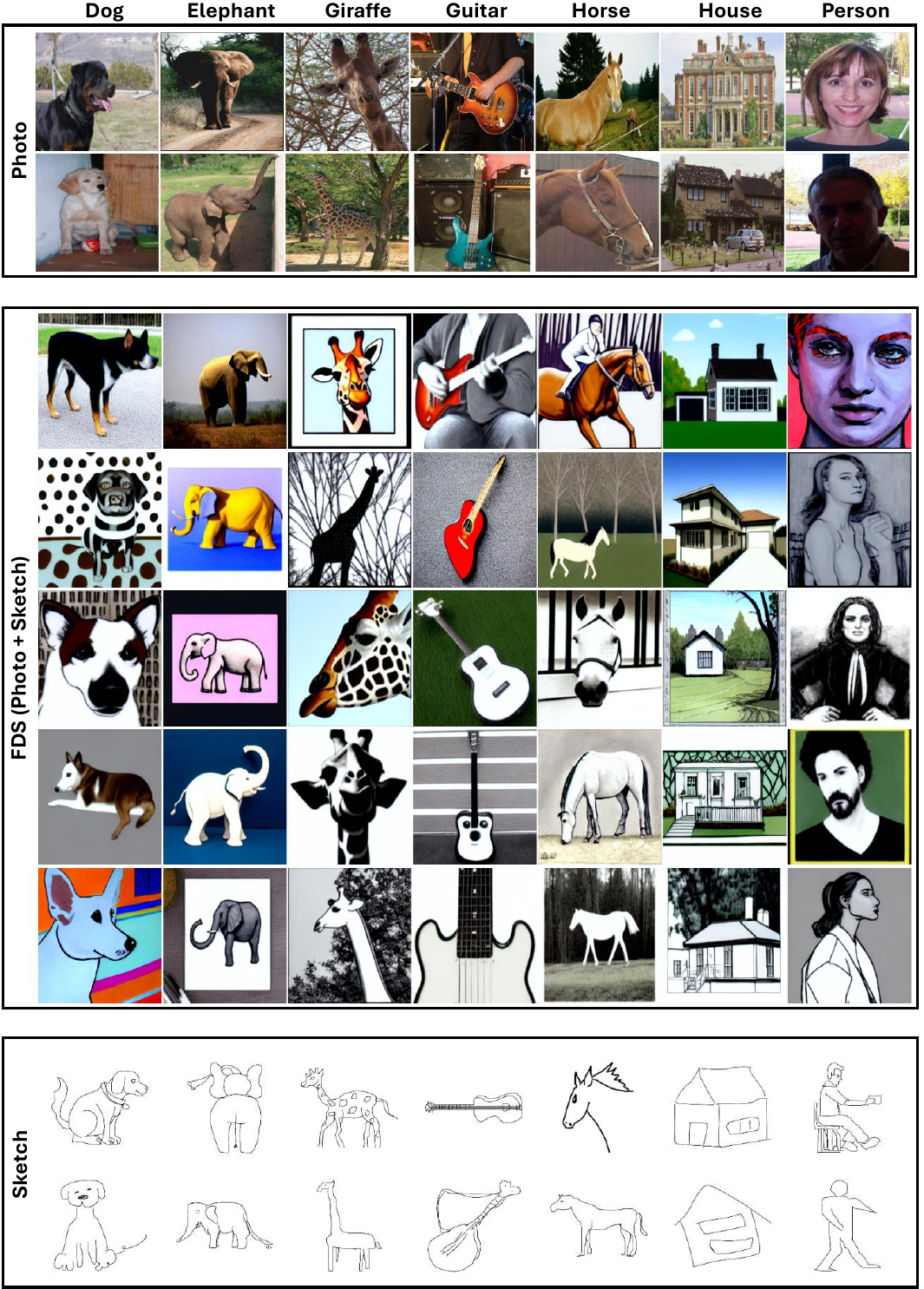}
  \caption{Visual comparison of original "Photo" and "Sketch" samples from PACS with synthetic images generated by FDS (Photo + Sketch). The middle section demonstrates how FDS integrates the photorealistic details of the "Photo" domain with the elements of the "Sketch" domain.}
  \label{fig:fds_ps}
\end{figure*}

\mypar{Visual Comparisons.}
This section visually compares the original images from the PACS dataset with the synthetic images generated by our FDS method, highlighting the ability of FDS to interpolate between domains. We provide examples for each pair of source domains used in training: "Art," "Photo," and "Sketch.". The visual comparisions are illustrated in Figures~\ref{fig:fds_ap}, \ref{fig:fds_sa}, and \ref{fig:fds_ps}. Each figure contains three sections: the first section shows samples from one original PACS domain, the middle section contains FDS-generated images combining the two selected domains, and the final section shows samples from the other original domain. These visual comparisons show that the FDS-generated images effectively blend domain-specific features, offering new pseudo-domain that can enrich the training set and enhance model generalization.

%%%%%%%%%%%%%%%%%%%%%%%%%%%%%%%%%%%%%%%%%%%%%%%%%%%%%%%%%%%%%%%%%%%5
\section{Oracle Results}
In addition to leave-one-out setting, where the validation set is selected from the training domains, some studies also report the results of oracle (test-domain validation set). This can be particularly useful for understanding the potential of a method when domain knowledge is available. In this section, we compare our method (FDS+ERM) with the state-of-the-art results, as shown in Table~\ref{tab:oracle_sota}. It is important to note that no Weight Averaging (WA) methods reported their oracle results within the DomainBed framework for a fair comparison. Therefore, we only train and report our ERM results here. Our proposed method, FDS+ERM, demonstrates superior performance across multiple benchmarks. Specifically, it achieves an average accuracy of 81.2\%, outperforming all other methods. On the PACS dataset, FDS+ERM attains the highest accuracy of 89.7\%, with significant improvements in the VLCS and OfficeHome datasets as well, achieving accuracies of 82.0\% and 71.8\% respectively. In addition to leave-one-out setting, these results also highlight the effectiveness of our approach in enhancing the performance under the oracle setting.

\begin{table*}[!t]
    \centering
    \small
    \tabcolsep=0.05cm
    \renewcommand{\arraystretch}{1.15} % Increase space between rows
    \adjustbox{max width=\textwidth}{
    \begin{tabular}{ll|c|p{0.1cm}cp{0.1cm}cp{0.1cm}cp{0.1cm}|cp{0.1cm}c}
    \toprule
     & \textbf{Method} & \textbf{ Aug. }           && \textbf{PACS}   && \textbf{VLCS}             && \textbf{OfficeHome}          && \textbf{  ~Avg. }              \\
    \midrule
    \multirow{15}{*}{\rotatebox{90}{Standard Methods}}~ 
    & ERM (\emph{baseline})~\cite{Gulrajani2021InSO} & \xmark  && 86.7\ppm0.3              && 77.6\ppm0.3              && 66.4\ppm0.5            && 76.9\\
    & ERM (\emph{reproduced}) & \xmark  && 86.6\ppm0.8              && 79.8\ppm0.4             && 68.4\ppm0.3           && 78.3 \\
    & IRM~\cite{arjovsky2019invariant} & \xmark           && 84.5\ppm1.1              && 76.9\ppm0.6              && 63.0\ppm2.7            && 74.8 \\
    & GroupDRO~\cite{sagawa2019distributionally} & \xmark && 87.1\ppm0.1              && 77.4\ppm0.5              && 66.2\ppm0.6            && 76.9 \\
    & Mixup~\cite{yan2020improve} & \cmark             && 86.8\ppm0.3              && 78.1\ppm0.3              && 68.0\ppm0.2            && 77.6 \\
    & CORAL~\cite{sun2016deep} & \xmark                  && 87.1\ppm0.5              && 77.7\ppm0.2              && 68.4\ppm0.2            && 77.7 \\
    & MMD~\cite{li2018domain} & \xmark                   && 87.2\ppm0.1              && 77.9\ppm0.1              && 66.2\ppm0.3            && 77.1 \\
    & DANN~\cite{ganin2016domain} & \xmark                && 85.2\ppm0.2              &&  79.7\ppm0.5              && 65.3\ppm0.8            && 76.7 \\
    & SagNet~\cite{Nam_2021_CVPR} & \cmark              && 86.4\ppm0.4              && 77.6\ppm0.1              && 67.5\ppm0.2            && 77.2 \\
    & RSC~\cite{huang2020self} & \cmark                && 86.2\ppm0.5              && ---              && 66.5\ppm0.6            && --- \\
    & SelfReg~\cite{kim2021selfreg} & \cmark              && 86.7\ppm0.8             &&  78.2\ppm0.1              && 68.1\ppm0.3            && 77.7 \\
    & Fishr~\cite{rame2022fishr} & \xmark              && 85.8\ppm0.6             &&   78.2\ppm0.2             && 66.0\ppm2.9            && 76.7 \\
    & CDGA~\cite{hemati2023cross} & \cmark              && \underline{89.6\ppm0.3}             && \underline{80.9\ppm0.1}              &&  \underline{68.8\ppm0.3}            && \underline{79.3}  \\
    \rowcolor{gray!15} & \textbf{ERM + FDS (ours)} & \cmark   && \textbf{89.7\ppm0.8}               && \textbf{82.0\ppm0.1}             && \textbf{71.8\ppm0.9}           && \textbf{81.2}  \\
    \bottomrule
    \end{tabular}}
    \caption{Oracle (test-domain validation set) accuracy\,(\%) results on the PACS, VLCS, and OfficeHome benchmarks. "\textbf{Aug.}" indicates whether advanced augmentation or domain mixing techniques are used. The \textbf{best results} and \underline{second-best results} are highlighted.}
    \label{tab:oracle_sota}
\end{table*}

\section{Detailed Results}

Here we present the comprehensive tables containing all the detailed information that was summarized in the main paper. The leave-one-out performance (train-domain validation set) across different domains of PACS, VLCS, and OfficeHome datasets are detailed in the tables \ref{tab:pacs}, \ref{tab:vlcs}, and \ref{tab:oh}, respectively. Additionally, the oracle (test-domain validation set) accuracy results for the PACS, VLCS, and OfficeHome benchmarks are detailed in Table \ref{tab:pacs_oracle}, \ref{tab:vlcs_oracle}, and \ref{tab:oh_oracle}, respectively.

\begin{table*}[t]

    \centering
    \tabcolsep=0.2cm
    \adjustbox{max width=\textwidth}{
      \begin{tabular}{cl|c|cccc|c}
        \toprule
        & \multirow[b]{2}{*}{\bf Method} & \multirow[b]{2}{*}{\bf Aug.} & \multicolumn{5}{c}{\bf Target Domains} \\
        \cmidrule(l{4pt}r{4pt}){4-8}
         &  &  & \bf Art & \bf Cartoon & \bf Photo & \bf Sketch & \bf Avg.\\
        \midrule
        \multirow{20}{*}{\rotatebox{90}{Standard Methods}} & ERM (baseline)~\cite{Gulrajani2021InSO}    & \xmark &  84.7\ppm0.4 & 80.8\ppm0.6 & 97.2\ppm0.3 & 79.3\ppm1.0 & 85.5\ppm0.2 \\
         & ERM (reproduced)                           & \xmark & 86.9\ppm0.6 & 80.2\ppm0.7 & 96.6\ppm0.4 & 74.5\ppm2.9 & 84.3\ppm1.1 \\
         & IRM~\cite{arjovsky2019invariant}           & \xmark & 84.8\ppm1.3 & 76.4\ppm1.1 & 96.7\ppm0.6 & 76.1\ppm1.0 & 83.5\ppm0.8 \\
         & GroupDRO~\cite{sagawa2019distributionally} & \xmark & 83.5\ppm0.9 & 79.1\ppm0.6 & 96.7\ppm0.3 & 78.3\ppm2.0 & 84.4\ppm0.8 \\
         & Mixup~\cite{yan2020improve}                & \cmark & 86.1\ppm0.5 & 78.9\ppm0.8 & \textbf{97.6\ppm0.1} & 75.8\ppm1.8 & 84.6\ppm0.6 \\
         & CORAL~\cite{sun2016deep}                   & \xmark & 88.3\ppm0.2 & 80.0\ppm0.5 & 97.5\ppm0.3 & 78.8\ppm1.3 & 86.2\ppm0.3 \\
         & MMD~\cite{li2018domain}                    & \xmark & 86.1\ppm1.4 & 79.4\ppm0.9 & 96.6\ppm0.2 & 76.5\ppm0.5 & 84.6\ppm0.5 \\
         & DANN~\cite{ganin2016domain}                & \xmark & 86.4\ppm0.8 & 77.4\ppm0.8 & 97.3\ppm0.4 & 73.5\ppm2.3 & 83.6\ppm0.4 \\
         & MLDG~\cite{li2018learning}                 & \xmark & 85.5\ppm1.4 & 80.1\ppm1.7 & 97.4\ppm0.3 & 76.6\ppm1.1 & 84.9\ppm1.1 \\
         & VREx~\cite{krueger2021out}                 & \xmark & 86.0\ppm1.6 & 79.1\ppm0.6 & 96.9\ppm0.5 & 77.7\ppm1.7 & 84.9\ppm1.1 \\
         & ARM~\cite{zhang2021adaptive}               & \xmark & 86.8\ppm0.6 & 76.8\ppm0.5 & 97.4\ppm0.3 & 79.3\ppm1.2 & 85.1\ppm0.6 \\
         & SagNet~\cite{Nam_2021_CVPR}                & \cmark & 87.4\ppm1.0 & 80.7\ppm0.6 & 97.1\ppm0.1 & 80.0\ppm0.4 & 86.3\ppm0.2 \\
         & RSC~\cite{huang2020self}                   & \cmark & 85.4\ppm0.8 & 79.7\ppm1.8 & \underline{97.6\ppm0.3} & 78.2\ppm1.2 & 85.2\ppm0.9 \\
         & Mixstyle~\cite{zhou2021domain}             & \cmark & 86.8\ppm0.5 & 79.0\ppm1.4 & 96.6\ppm0.1 & 78.5\ppm2.3 & 85.2\ppm0.3 \\
         & mDSDI~\cite{bui2021exploiting}             & \xmark & 87.7\ppm0.4 & 80.4\ppm0.7 & 98.1\ppm0.3 & 78.4\ppm1.2 & 86.2\ppm0.2 \\
         & SelfReg~\cite{kim2021selfreg}              & \cmark & 87.9\ppm1.0  & 79.4\ppm1.4 & 96.8\ppm0.7 & 78.3\ppm1.2 & 85.6\ppm0.4 \\
         & Fishr~\cite{rame2022fishr}                 & \xmark & 88.4\ppm0.2 & 78.7\ppm0.7 & 97.0\ppm0.1 & 77.8\ppm2.0 & 85.5\ppm0.5 \\
         & DCAug~\cite{aminbeidokhti2024domain}       & \cmark & 88.5\ppm0.8 & 78.8\ppm1.5 & 96.3\ppm0.1 & 80.8\ppm0.5 & 86.1\ppm0.7 \\
         & DomainDiff~\cite{miao2024domaindiff}       & \cmark & 84.9\ppm1.6  & \underline{82.9\ppm0.0} &  95.5\ppm0.0 & 79.0\ppm0.9 & 85.6\ppm0.6 \\
         & DSI~\cite{yu2023distribution}       & \cmark & 84.6\ppm2.4 & 81.4\ppm1.6  & 96.8\ppm0.5  & 82.5\ppm1.0  & 86.9\ppm1.4   \\
         & CDGA~\cite{hemati2023cross}                & \cmark & \underline{89.1\ppm1.0} & 82.5\ppm0.5 & 97.4\ppm0.2 & \textbf{84.8\ppm0.9} & \underline{88.5\ppm0.5}  \\
        \rowcolor{gray!15}
         & \textbf{ERM + FDS (ours)}                           & \cmark & \textbf{90.7\ppm0.9} & \textbf{84.2\ppm0.6} & 97.2\ppm0.1 & \underline{83.0\ppm0.4} & \textbf{88.8\ppm0.1} \\
        \midrule
        \multirow{7}{*}{\rotatebox{90}{WA Methods}} & SWAD (baseline)~\cite{cha2021swad}                         & \xmark & 89.3\ppm0.2  & 83.4\ppm0.6 & 97.3\ppm0.3 & 82.5\ppm0.5 & 88.1\ppm0.1 \\
        & SWAD (reproduced)                         & \xmark & 89.5\ppm0.2 & \underline{83.7\ppm0.4} & 97.3\ppm0.2 & 82.1\ppm0.1 & 88.1\ppm0.4 \\
        & SelfReg SWA~\cite{kim2021selfreg}         & \cmark & 85.9\ppm0.6 & 81.9\ppm0.4 & 96.8\ppm0.1 & 81.4\ppm0.6 & 86.5\ppm0.3 \\
        & DNA~\cite{chu2022dna}                     & \xmark & 89.8\ppm0.2 & 83.4\ppm0.4 &  97.7\ppm0.1  & 82.6\ppm0.2 & 88.4\ppm0.1 \\
        & DiWA~\cite{rame2022diverse}               & \cmark & \underline{90.1\ppm0.6} & 83.3\ppm0.6 &  \textbf{98.2\ppm0.1} & 83.4\ppm0.4 &  \underline{88.8\ppm0.4} \\
        & TeachDCAug~\cite{aminbeidokhti2024domain} & \cmark & 89.6\ppm0.0 & 81.8\ppm0.5 & 97.7\ppm0.0  & \underline{84.5\ppm0.2} & 88.4\ppm0.2 \\
        \rowcolor{gray!15}
         & \textbf{SWAD + FDS (ours)}                         & \cmark & \textbf{91.8\ppm0.3} & \textbf{86.0\ppm0.8} & \underline{98.1\ppm0.2} & \textbf{86.1\ppm0.1} & \textbf{90.5\ppm0.3} \\
      \bottomrule
    \end{tabular}
}
    \caption{Leave-one-out accuracy\,(\%) results on the PACS dataset. "\textbf{Aug.}" indicates whether advanced augmentation or domain mixing techniques are used. The \textbf{best results} and \underline{second-best results} are highlighted.}
    \label{tab:pacs}
\end{table*}

\begin{table*}[t]

    \centering
    \tabcolsep=0.2cm
    \adjustbox{max width=\textwidth}{
      \begin{tabular}{cl|c|cccc|c}
        \toprule
        & \multirow[b]{2}{*}{\bf Method} & \multirow[b]{2}{*}{\bf Aug.} & \multicolumn{5}{c}{\bf Target Domains} \\
        \cmidrule(l{4pt}r{4pt}){4-8}
         &  &  & \bf Caltech101 & \bf LabelMe & \bf SUN09 & \bf VOC2007 & \bf Avg.\\
        \midrule
        \multirow{20}{*}{\rotatebox{90}{Standard Methods}} & ERM (baseline)~\cite{Gulrajani2021InSO}    & \xmark & 97.7\ppm0.4 & 64.3\ppm0.9 & 73.4\ppm0.5 & 74.6\ppm1.3 & 77.5\ppm0.4 \\
         & ERM (reproduced)                          & \xmark & 96.9\ppm1.4 & 64.1\ppm1.4 & 71.1\ppm1.5 & 72.8\ppm0.9 & 76.2\ppm1.1 \\
         & IRM~\cite{arjovsky2019invariant}                                       & \xmark & 98.6\ppm0.1 & 64.9\ppm0.9 & 73.4\ppm0.6 & 77.3\ppm0.9 & 78.5\ppm0.5 \\
         & GroupDRO~\cite{sagawa2019distributionally}                                  & \xmark & 97.3\ppm0.3 & 63.4\ppm0.9 & 69.5\ppm0.8 & 76.7\ppm0.7 & 76.7\ppm0.6 \\
         & Mixup~\cite{yan2020improve}                                     & \cmark & 98.3\ppm0.6 & 64.8\ppm1.0 & 72.1\ppm0.5 & 74.3\ppm0.8 & 77.4\ppm0.6 \\
         & CORAL~\cite{sun2016deep}                                    & \xmark & 98.3\ppm0.1 & 66.1\ppm1.2 & 73.4\ppm0.3 & 77.5\ppm1.2 & 78.8\ppm0.6 \\
         & MMD~\cite{li2018domain}                                       & \xmark & 97.7\ppm0.1 & 64.0\ppm1.1 & 72.8\ppm0.2 & 75.3\ppm3.3 & 77.5\ppm0.9 \\
         & DANN~\cite{ganin2016domain}                                      & \xmark & \textbf{99.0\ppm0.3} & 65.1\ppm1.4 & 73.1\ppm0.3 & 77.2\ppm0.6 & 78.6\ppm0.4 \\
         & MLDG~\cite{li2018learning}                                      & \xmark & 97.4\ppm0.2 & 65.2\ppm0.7 & 71.0\ppm1.4 & 75.3\ppm1.0 & 77.2\ppm0.8 \\
         & VREx~\cite{krueger2021out}                                       & \xmark & 98.4\ppm0.3 & 64.4\ppm1.4 & 74.1\ppm0.4 & 76.2\ppm1.3 & 78.3\ppm0.8 \\
         & ARM~\cite{zhang2021adaptive}                                        & \xmark & 98.7\ppm0.2 & 63.6\ppm0.7 & 71.3\ppm1.2 & 76.7\ppm0.6 & 77.6\ppm0.6 \\
         & SagNet~\cite{Nam_2021_CVPR}                                    & \cmark & 97.9\ppm0.4 & 64.5\ppm0.5 & 71.4\ppm1.3 & 77.5\ppm0.5 & 77.8\ppm0.5 \\
         & RSC~\cite{huang2020self}                                       & \cmark & 97.9\ppm0.1 & 62.5\ppm0.7 & 72.3\ppm1.2 & 75.6\ppm0.8 & 77.1\ppm0.5 \\
         & Mixstyle~\cite{zhou2021domain}                                  & \cmark & 98.6\ppm0.3 & 64.5\ppm1.1 & 72.6\ppm0.5 & 75.7\ppm1.7 & 77.9\ppm0.5 \\
         & mDSDI~\cite{bui2021exploiting}                                     & \xmark & 97.6\ppm0.1 & \underline{66.4\ppm0.4} & 74.0\ppm0.6 & \underline{77.8\ppm0.7} & 79.0\ppm0.3 \\
         & SelfReg~\cite{kim2021selfreg}                                   & \cmark & 96.7\ppm0.4 & 65.2\ppm1.2 & 73.1\ppm1.3 & 76.2\ppm0.7 & 77.8\ppm0.9 \\
         & Fishr~\cite{rame2022fishr}                                     & \xmark & \underline{98.9\ppm0.3} & 64.0\ppm0.5 & 71.5\ppm0.2 & 76.8\ppm0.7 & 77.8\ppm0.5 \\
         & DCAug~\cite{aminbeidokhti2024domain}                                     & \cmark & 98.3\ppm0.1 & 64.2\ppm0.4 & \underline{74.4\ppm0.6} & 77.5\ppm0.3 & 78.6\ppm0.5 \\
         & CDGA~\cite{hemati2023cross}                                       & \cmark & 96.3\ppm0.7 & \textbf{75.7\ppm1.0} & 72.8\ppm1.3 & 73.7\ppm1.3 & \underline{79.6\ppm0.9} \\
        \rowcolor{gray!15}
         & \textbf{ERM + FDS (ours)}                           & \cmark & 98.8\ppm0.3 & 65.6\ppm0.9 & \textbf{75.5\ppm0.9} & \textbf{79.3\ppm1.8} & \textbf{79.8\ppm0.5}  \\
        \midrule
        \multirow{7}{*}{\rotatebox{90}{WA Methods}} & SWAD (baseline)~\cite{cha2021swad}                         & \xmark & \underline{98.8\ppm0.1} & 63.3\ppm0.3 & 75.3\ppm0.5 & 79.2\ppm0.6 & \underline{79.1\ppm0.1} \\
        & SWAD (reproduced)                         & \xmark & 98.7\ppm0.2 & \textbf{63.9\ppm0.3} & 74.3\ppm1.1 & 78.6\ppm0.6 & 78.9\ppm0.5 \\
        & SelfReg SWA~\cite{kim2021selfreg}                               & \cmark & 97.4\ppm0.4 & 63.5\ppm0.3 & 72.6\ppm0.1 & 76.7\ppm0.7 & 77.5\ppm0.0 \\
         & DNA~\cite{chu2022dna}                                       & \xmark & \underline{98.8\ppm0.1} & 63.6\ppm0.2 & 74.1\ppm0.1 & \underline{79.5\ppm0.4} & 79.0\ppm0.5 \\
         & DiWA~\cite{rame2022diverse}                                      & \cmark & 98.4\ppm0.1 & 63.4\ppm0.1 & 75.5\ppm0.3 & 78.9\ppm0.6 & 79.1\ppm0.2 \\
        & TeachDCAug~\cite{aminbeidokhti2024domain}                                & \cmark & 98.5\ppm0.1 & \underline{63.7\ppm0.3} & \underline{75.6\ppm0.5} & 77.0\ppm0.7 & 78.7\ppm0.5 \\
        \rowcolor{gray!15}
         & \textbf{SWAD + FDS (ours)}                         & \cmark & \textbf{99.5\ppm0.2} & 62.9\ppm0.2 & \textbf{76.9\ppm0.4} & \textbf{79.6\ppm1.3} & \textbf{79.7\ppm0.5} \\
      \bottomrule
    \end{tabular}
}
    \caption{Leave-one-out accuracy\,(\%) results on the VLCS dataset. "\textbf{Aug.}" indicates whether advanced augmentation or domain mixing techniques are used. The \textbf{best results} and \underline{second-best results} are highlighted.}
    \label{tab:vlcs}
\end{table*}

\begin{table*}[t]

    \centering
    \tabcolsep=0.2cm
    \adjustbox{max width=\textwidth}{
      \begin{tabular}{cl|c|cccc|c}
        \toprule
        & \multirow[b]{2}{*}{\bf Method} & \multirow[b]{2}{*}{\bf Aug.} & \multicolumn{5}{c}{\bf Target Domains} \\
        \cmidrule(l{4pt}r{4pt}){4-8}
         &  &  & \bf Art & \bf Clipart & \bf Product & \bf Real World & \bf Avg.\\
        \midrule
        \multirow{20}{*}{\rotatebox{90}{Standard Methods}} & ERM (baseline)~\cite{Gulrajani2021InSO}  & \xmark & 61.3\ppm0.7 & 52.4\ppm0.3 & 75.8\ppm0.1 & 76.6\ppm0.3 & 66.5\ppm0.3 \\
         & ERM (reproduced)                             & \xmark & 59.5\ppm2.1 & 51.3\ppm1.3 & 73.8\ppm0.8 & 73.8\ppm0.2 & 64.6\ppm1.1 \\
         & IRM~\cite{arjovsky2019invariant}             & \xmark & 58.9\ppm2.3 & 52.2\ppm1.6 & 72.1\ppm2.9 & 74.0\ppm2.5 & 64.3\ppm2.2 \\
         & GroupDRO~\cite{sagawa2019distributionally}   & \xmark & 60.4\ppm0.7 & 52.7\ppm1.0 & 75.0\ppm0.7 & 76.0\ppm0.7 & 66.0\ppm0.7 \\
         & Mixup~\cite{yan2020improve}                  & \cmark & 62.4\ppm0.8 & 54.8\ppm0.6 & 76.9\ppm0.3 & 78.3\ppm0.2 & 68.1\ppm0.3 \\
         & CORAL~\cite{sun2016deep}                     & \xmark & \underline{65.3\ppm0.4} & 54.4\ppm0.5 & 76.5\ppm0.1 & 78.4\ppm0.5 & 68.7\ppm0.3 \\
         & MMD~\cite{li2018domain}                      & \xmark & 60.4\ppm0.2 & 53.3\ppm0.3 & 74.3\ppm0.1 & 77.4\ppm0.6 & 66.3\ppm0.1 \\
         & DANN~\cite{ganin2016domain}                  & \xmark & 59.9\ppm1.3 & 53.0\ppm0.3 & 73.6\ppm0.7 & 76.9\ppm0.5 & 65.9\ppm0.6 \\
         & MLDG~\cite{li2018learning}                   & \xmark & 61.5\ppm0.9 & 53.2\ppm0.6 & 75.0\ppm1.2 & 77.5\ppm0.4 & 66.8\ppm0.7 \\
         & VREx~\cite{krueger2021out}                   & \xmark & 60.7\ppm0.9 & 53.0\ppm0.9 & 75.3\ppm0.1 & 76.6\ppm0.5 & 66.4\ppm0.6 \\
         & ARM~\cite{zhang2021adaptive}                 & \xmark & 58.9\ppm0.8 & 51.0\ppm0.5 & 74.1\ppm0.1 & 75.2\ppm0.3 & 64.8\ppm0.4 \\
         & SagNet~\cite{Nam_2021_CVPR}                  & \cmark & 63.4\ppm0.2 & 54.8\ppm0.4 & 75.8\ppm0.4 & 78.3\ppm0.3 & 68.1\ppm0.1 \\
         & RSC~\cite{huang2020self}                     & \cmark & 60.7\ppm1.4 & 51.4\ppm0.3 & 74.8\ppm1.1 & 75.1\ppm1.3 & 65.5\ppm0.9 \\
         & Mixstyle~\cite{zhou2021domain}               & \cmark & 51.1\ppm0.3 & 53.2\ppm0.4 & 68.2\ppm0.7 & 69.2\ppm0.6 & 60.4\ppm0.3 \\
         & mDSDI~\cite{bui2021exploiting}               & \xmark & \textbf{68.1\ppm0.3} & 52.1\ppm0.4 & 76.0\ppm0.2 & 80.4\ppm0.2 & \underline{69.2\ppm0.4} \\
         & SelfReg~\cite{kim2021selfreg}                & \cmark & 63.6\ppm1.4 & 53.1\ppm1.0 & 76.9\ppm0.4 & 78.1\ppm0.4 & 67.9\ppm0.7 \\
         & Fishr~\cite{rame2022fishr}                   & \xmark & 62.4\ppm0.5 & 54.4\ppm0.4 & 76.2\ppm0.5 & 78.3\ppm0.1 & 67.8\ppm0.5 \\
         & DCAug~\cite{aminbeidokhti2024domain}         & \cmark & 61.8\ppm0.6 & \underline{55.4\ppm0.6} & 77.1\ppm0.3 & 78.9\ppm0.3 & 68.3\ppm0.4 \\
         & DomainDiff~\cite{miao2024domaindiff}         & \cmark & 57.6\ppm0.4  & 49.2\ppm0.6 & 73.0\ppm0.6 & 75.2\ppm0.9 & 63.7\ppm0.6 \\
         & CDGA~\cite{hemati2023cross}                  & \cmark & 60.1\ppm1.4 & 54.2\ppm0.5 & \underline{78.2\ppm0.6} & \underline{80.4\ppm0.1} & 68.2\ppm0.6 \\
         & \textbf{ERM + FDS (ours)}                             & \cmark & 64.6\ppm0.2 & \textbf{57.7\ppm0.1} & \textbf{80.2\ppm0.5} & \textbf{82.0\ppm0.4} & \textbf{71.1\ppm0.1} \\
        \midrule
        \multirow{7}{*}{\rotatebox{90}{WA Methods}}     & SWAD (baseline)~\cite{cha2021swad}  & \xmark & 66.1\ppm0.4 & 57.7\ppm0.4 & 78.4\ppm0.1 & 80.2\ppm0.2 & 70.6\ppm0.2 \\
         & SWAD (reproduced)                            & \xmark & 65.9\ppm0.9 & 56.8\ppm0.4 & 78.8\ppm0.3 & 80.0\ppm0.2 & 70.3\ppm0.4 \\
         & SelfReg SWA~\cite{kim2021selfreg}            & \cmark & 64.9\ppm0.8 & 55.4\ppm0.6 & 78.4\ppm0.2 & 78.8\ppm0.1 & 69.4\ppm0.2 \\
         & DNA~\cite{chu2022dna}                        & \xmark & \textbf{67.7\ppm0.2} & 57.7\ppm0.3 & 78.9\ppm0.2 & \underline{80.5\ppm0.2} & \underline{71.2\ppm0.1} \\
         & DiWA~\cite{rame2022diverse}                  & \cmark & \underline{67.3\ppm0.2} & \underline{57.9\ppm0.2} & \underline{79.0\ppm0.2} & 79.9\ppm0.1 & 71.0\ppm0.1 \\
         & TeachDCAug~\cite{aminbeidokhti2024domain}    & \cmark & 66.2\ppm0.2 & 57.0\ppm0.3 & 78.3\ppm0.1 & 80.1\ppm0.0 & 70.4\ppm0.2 \\
         & \textbf{SWAD + FDS (ours)}                            & \cmark & 67.3\ppm0.8 & \textbf{60.5\ppm0.5} & \textbf{82.6\ppm0.1} & \textbf{83.6\ppm0.3} & \textbf{73.5\ppm0.4} \\

      \bottomrule
    \end{tabular}
}
    \caption{Leave-one-out accuracy\,(\%) results on the OfficeHome dataset. "\textbf{Aug.}" indicates whether advanced augmentation or domain mixing techniques are used. The \textbf{best results} and \underline{second-best results} are highlighted.}
    \label{tab:oh}
\end{table*}

\begin{table*}[t]

    \centering
    \tabcolsep=0.2cm
    \adjustbox{max width=\textwidth}{
      \begin{tabular}{cl|c|cccc|c}
        \toprule
        & \multirow[b]{2}{*}{\bf Method} & \multirow[b]{2}{*}{\bf Aug.} & \multicolumn{5}{c}{\bf Target Domains} \\
        \cmidrule(l{4pt}r{4pt}){4-8}
         &  &  & \bf Art & \bf Cartoon & \bf Photo & \bf Sketch & \bf Avg.\\
        \midrule
        \multirow{14}{*}{\rotatebox{90}{Standard Methods}} & ERM (baseline)~\cite{Gulrajani2021InSO}    & \xmark & 86.5\ppm1.0 & 81.3\ppm0.6 & 96.2\ppm0.3 & 82.7\ppm1.1 & 86.7\ppm0.8 \\
         & ERM (reproduced)                    & \xmark & 88.6\ppm0.9 & 80.9\ppm1.9 & \textbf{98.4\ppm0.4} & 78.4\ppm1.2 & 86.6\ppm1.0 \\
         & IRM                                 & \xmark & 84.2\ppm0.9 & 79.7\ppm1.5 & 95.9\ppm0.4 & 78.3\ppm2.1 & 84.5\ppm1.2 \\
         & GroupDRO                            & \xmark & 87.5\ppm0.5 & 82.9\ppm0.6 & 97.1\ppm0.3 & 81.1\ppm1.2 & 87.2\ppm0.7 \\
         & Mixup                               & \cmark & 87.5\ppm0.4 & 81.6\ppm0.7 & 97.4\ppm0.2 & 80.8\ppm0.9 & 86.8\ppm0.6 \\
         & CORAL                               & \xmark & 86.6\ppm0.8 & 81.8\ppm0.9 & 97.1\ppm0.5 & 82.7\ppm0.6 & 87.1\ppm0.7 \\
         & MMD                                 & \xmark & 88.1\ppm0.8 & 82.6\ppm0.7 & 97.1\ppm0.5 & 81.2\ppm1.2 & 87.3\ppm0.8 \\
         & DANN                                & \xmark & 87.0\ppm0.4 & 80.3\ppm0.6 & 96.8\ppm0.3 & 76.9\ppm1.1 & 85.3\ppm0.6 \\
         & SagNet                              & \cmark & 87.4\ppm0.5 & 81.2\ppm1.2 & 96.3\ppm0.8 & 80.7\ppm1.1 & 86.4\ppm0.9 \\
         & RSC                                 & \cmark & 86.0\ppm0.7 & 81.8\ppm0.9 & 96.8\ppm0.7 & 80.4\ppm0.5 & 86.3\ppm0.7 \\
         & Fishr                               & \xmark & 87.9\ppm0.6 & 80.8\ppm0.5 & \underline{97.9\ppm0.4} & 81.1\ppm0.8 & 86.9\ppm0.6 \\
         & SelfReg                             & \cmark & 87.9\ppm0.5 & 80.6\ppm1.1 & 97.1\ppm0.4 & 81.1\ppm1.3 & 86.7\ppm0.8 \\
         & CDGA                                & \cmark & \underline{89.6\ppm0.8} & \textbf{85.3\ppm0.7} & 97.3\ppm0.3 & \textbf{86.2\ppm0.5} & \underline{89.6\ppm0.6} \\
        \rowcolor{gray!15}
         & \textbf{ERM + FDS (ours)}           & \cmark & \textbf{91.1\ppm0.3} & \underline{84.9\ppm0.7} & 97.3\ppm0.5 & \underline{85.6\ppm2.3} & \textbf{89.7\ppm0.8} \\
        \bottomrule
    \end{tabular}
}
    \caption{Oracle (test-domain validation set) accuracy\,(\%) results on the PACS dataset. "\textbf{Aug.}" indicates whether advanced augmentation or domain mixing techniques are used. The \textbf{best results} and \underline{second-best results} are highlighted.}
    \label{tab:pacs_oracle}
\end{table*}

\begin{table*}[t]

    \centering
    \tabcolsep=0.2cm
    \adjustbox{max width=\textwidth}{
      \begin{tabular}{cl|c|cccc|c}
        \toprule
        & \multirow[b]{2}{*}{\bf Method} & \multirow[b]{2}{*}{\bf Aug.} & \multicolumn{5}{c}{\bf Target Domains} \\
        \cmidrule(l{4pt}r{4pt}){4-8}
         &  &  & \bf Caltech101 & \bf LabelMe & \bf SUN09 & \bf VOC2007 & \bf Avg.\\
        \midrule
        \multirow{14}{*}{\rotatebox{90}{Standard Methods}} & ERM (baseline)~\cite{Gulrajani2021InSO}    & \xmark & 97.6\ppm0.3 & 67.9\ppm0.7 & 70.9\ppm0.2 & 74.0\ppm0.6 & 77.6\ppm0.5 \\
         & ERM (reproduced)                          & \xmark & 98.6\ppm0.2 & 68.6\ppm0.7 & 73.6\ppm1.7 & 78.6\ppm1.2 & 79.8\ppm0.4 \\
         & IRM                                       & \xmark & 97.3\ppm0.2 & 66.7\ppm0.1 & 71.0\ppm2.3 & 72.8\ppm0.4 & 77.0\ppm0.8 \\
         & GroupDRO                                  & \xmark & 97.7\ppm0.2 & 65.9\ppm0.2 & 72.8\ppm0.8 & 73.4\ppm1.3 & 77.5\ppm0.6 \\
         & Mixup                                     & \cmark & 97.8\ppm0.4 & 67.2\ppm0.4 & 71.5\ppm0.2 & 75.7\ppm0.6 & 78.1\ppm0.4 \\
         & CORAL                                     & \xmark & 97.3\ppm0.2 & 67.5\ppm0.6 & 71.6\ppm0.6 & 74.5\ppm0.0 & 77.7\ppm0.4 \\
         & MMD                                       & \xmark & 98.8\ppm0.0 & 66.4\ppm0.4 & 70.8\ppm0.5 & 75.6\ppm0.4 & 77.9\ppm0.3 \\
         & DANN                                      & \xmark & \underline{9.0\ppm0.2} & 66.3\ppm1.2 & 73.4\ppm1.4 & \underline{80.1\ppm0.5} & 79.7\ppm0.8 \\
         & SagNet                                    & \cmark & 97.4\ppm0.3 & 66.4\ppm0.4 & 71.6\ppm0.1 & 75.0\ppm0.8 & 77.6\ppm0.4 \\
         & RSC                                       & \cmark & 98.0\ppm0.4 & 67.2\ppm0.3 & 70.3\ppm1.3 & 75.6\ppm0.4 & 77.8\ppm0.6 \\
         & Fishr                                     & \xmark & 97.6\ppm0.7 & 67.3\ppm0.5 & 72.2\ppm0.9 & 75.7\ppm0.3 & 78.2\ppm0.6 \\
         & SelfReg                                   & \cmark & 98.2\ppm0.3 & 63.9\ppm0.8 & 72.2\ppm0.1 & 75.5\ppm0.4 & 77.5\ppm0.2 \\
         & CDGA                                      & \cmark & 96.6\ppm0.7 & \textbf{75.5\ppm1.9} & \underline{73.6\ppm1.1} & 77.8\ppm1.0 & \underline{80.9\ppm1.2} \\
         \rowcolor{gray!15}
         & \textbf{ERM + FDS (ours)}                          & \cmark & \textbf{99.5\ppm0.1} & \underline{68.7\ppm0.3} & \textbf{77.4\ppm0.7} & \textbf{82.6\ppm0.1} & \textbf{82.0\ppm0.1} \\
        \bottomrule
    \end{tabular}
}
    \caption{Oracle (test-domain validation set) accuracy\,(\%) results on the VLCS dataset. "\textbf{Aug.}" indicates whether advanced augmentation or domain mixing techniques are used. The \textbf{best results} and \underline{second-best results} are highlighted.}
    \label{tab:vlcs_oracle}
\end{table*}

\begin{table*}[t]

    \centering
    \tabcolsep=0.2cm
    \adjustbox{max width=\textwidth}{
      \begin{tabular}{cl|c|cccc|c}
        \toprule
        & \multirow[b]{2}{*}{\bf Method} & \multirow[b]{2}{*}{\bf Aug.} & \multicolumn{5}{c}{\bf Target Domains} \\
        \cmidrule(l{4pt}r{4pt}){4-8}
         &  &  & \bf Art & \bf Clipart & \bf Product & \bf Real World & \bf Avg.\\
        \midrule
        \multirow{14}{*}{\rotatebox{90}{Standard Methods}} & ERM (baseline)~\cite{Gulrajani2021InSO}    & \xmark & 61.7\ppm0.7 & 53.4\ppm0.3 & 74.1\ppm0.4 & 76.2\ppm0.6 & 66.4\ppm0.5 \\
         & ERM (reproduced)                          & \xmark & 64.0\ppm0.9 & 53.7\ppm1.1 & 77.1\ppm0.3 & 78.8\ppm0.4 & 68.4\ppm0.3 \\
         & IRM                                       & \xmark & 56.4\ppm3.2 & 51.2\ppm2.3 & 71.7\ppm2.7 & 72.7\ppm2.7 & 63.0\ppm2.7 \\
         & GroupDRO                                  & \xmark & 60.5\ppm1.6 & 53.1\ppm0.3 & 75.5\ppm0.3 & 75.9\ppm0.7 & 66.3\ppm0.7 \\
         & Mixup                                     & \cmark & 63.5\ppm0.2 & 54.6\ppm0.4 & 76.0\ppm0.3 & 78.0\ppm0.7 & 68.0\ppm0.4 \\
         & CORAL                                     & \xmark & \underline{64.8\ppm0.8} & 54.1\ppm0.9 & 76.5\ppm0.4 & 78.2\ppm0.4 & 68.4\ppm0.6 \\
         & MMD                                       & \xmark & 60.4\ppm1.0 & 53.4\ppm0.5 & 74.9\ppm0.1 & 76.1\ppm0.7 & 66.2\ppm0.6 \\
         & DANN                                      & \xmark & 60.6\ppm1.4 & 51.8\ppm0.7 & 73.4\ppm0.5 & 75.5\ppm0.9 & 65.3\ppm0.9 \\
         & SagNet                                    & \cmark & 62.7\ppm0.5 & 53.6\ppm0.5 & 76.0\ppm0.3 & 77.8\ppm0.1 & 67.5\ppm0.4 \\
         & RSC                                       & \cmark & 61.7\ppm0.8 & 53.0\ppm0.9 & 74.8\ppm0.8 & 76.3\ppm0.5 & 66.5\ppm0.8 \\
         & Fishr                                     & \xmark & 63.4\ppm0.8 & 54.2\ppm0.3 & 76.4\ppm0.3 & 78.5\ppm0.2 & 68.1\ppm0.4 \\
         & SelfReg                                   & \cmark & 64.2\ppm0.6 & 53.6\ppm0.7 & 76.7\ppm0.3 & 77.9\ppm0.5 & 68.1\ppm0.3 \\
         & CDGA                                      & \cmark & 61.1\ppm1.1 & \underline{55.9\ppm1.0} & \underline{78.2\ppm0.8} & \underline{79.8\ppm0.2} & \underline{68.8\ppm0.8} \\
         \rowcolor{gray!15}
         & \textbf{ERM + FDS (ours)}                          & \cmark & \textbf{65.3\ppm0.8} & \textbf{58.4\ppm0.8} & \textbf{81.2\ppm0.2} & \textbf{82.4\ppm0.6} & \textbf{71.8\ppm0.9} \\

        \bottomrule
    \end{tabular}
}
    \caption{Oracle (test-domain validation set) accuracy\,(\%) results on the OfficeHome dataset. "\textbf{Aug.}" indicates whether advanced augmentation or domain mixing techniques are used. The \textbf{best results} and \underline{second-best results} are highlighted.}
    \label{tab:oh_oracle}
\end{table*}

% \section{Additional Results}
% Oracle
\clearpage 

%%%%%%%%%%%%%%%%%%% FDS dataset

%%%%%%%%%%%%%%%%%%%%%%%%%%%%%%%%%%%%
%Bibliography
\clearpage 
\bibliographystyle{unsrt}  
\bibliography{references}

\end{document}